\documentclass{article}
\usepackage{arxiv}
\usepackage[utf8]{inputenc}
\usepackage[pdftex]{graphicx}
\usepackage{booktabs}
\usepackage{caption} 
\usepackage{subcaption}
\usepackage{algorithm}
\usepackage{algorithmic}
\usepackage{amsmath}
\usepackage[dvipsnames]{xcolor}
\usepackage{comment}
\usepackage{subfiles}
\usepackage{natbib}
\usepackage{multirow}

\title{A Hardware-Aware Framework for Accelerating Neural Architecture Search Across Modalities}

\author{
    Daniel Cummings \\
    \small{Intel Labs, Intel Corporation} \\
    \small{daniel.cummings@intel.com} \\
    \And
    Anthony Sarah \\
    \small{Intel Labs, Intel Corporation} \\
    \small{anthony.sarah@intel.com} \\
    \And
    Sharath Nittur Sridhar \\
    \small{Intel Labs, Intel Corporation} \\
    \small{sharath.nittur.sridhar@intel.com} \\
    \And
    Maciej Szankin \\
    \small{Intel Labs, Intel Corporation} \\
    \small{maciej.szankin@intel.com} \\
    \And
    Juan Pablo Munoz \\
    \small{Intel Labs, Intel Corporation} \\
    \small{pablo.munoz@intel.com} \\
    \And
    Sairam Sundaresan \\
    \small{Intel Labs, Intel Corporation} \\
    \small{sairam.sundaresan@intel.com} \\
}

\date{}

\begin{document}

\maketitle

\begin{abstract}
  Recent advances in Neural Architecture Search (NAS) such as one-shot NAS offer the ability to extract specialized hardware-aware sub-network configurations from a task-specific super-network. While considerable effort has been employed towards improving the first stage, namely, the training of the super-network, the search for derivative high-performing sub-networks is still under-explored. Popular methods decouple the super-network training from the sub-network search and use performance predictors to reduce the computational burden of searching on different hardware platforms. We propose a flexible search framework that automatically and efficiently finds optimal sub-networks that are optimized for different performance metrics and hardware configurations. Specifically, we show how evolutionary algorithms can be paired with lightly trained objective predictors in an iterative cycle to accelerate architecture search in a multi-objective setting for various modalities including machine translation and image classification. 
\end{abstract}

\section{Introduction}

Artificial intelligence researchers are continually pushing the state-of-the-art in deep learning model performance across many application domains. Neural architecture search (NAS) has become an increasingly popular technique to achieve these performance gains with results that often outperform hand-designed architectures. In many cases, the deep neural network (DNN) design and evaluation process is tied to the hardware platform available to the researcher at the time (e.g., GPU). Furthermore, the researcher may have only been interested in a single performance objective such as accuracy when evaluating the network. Therefore, the network is inherently optimized for a specific hardware platform and specific objective. However, users wanting to solve the same problem for which the network was designed may have different hardware platforms available and may be interested in multiple performance metrics (e.g., accuracy \textit{and} latency). The performance of the network provided by the researcher is then sub-optimal for these users. 

Unfortunately, optimizing the network for the user's hardware and performance objectives is a time-consuming effort requiring highly specialized knowledge. Network optimization is typically done manually with a great deal of in-depth understanding of the hardware platform since certain hardware characteristics (e.g., clock speed, number of logical processor cores, amount of RAM, architecture) will affect the optimization process. The optimization process is also affected by the characteristics of the input data to the DNN (e.g., batch size, image size). Finally, any change to the performance objectives (e.g., going from latency to power consumption), input data characteristics, hardware characteristics, or hardware platform (e.g., going from GPU to CPU) would require starting this expensive optimization process again. With this in mind, we adopt the weight-sharing super-network NAS approach that is well suited to the hardware-aware model optimization task. This NAS approach generates a highly-diverse set of architectural configurations (a.k.a. sub-networks) from a reference super-network architecture and allows us to explore an extremely large search space to find optimal models for a particular hardware and objective setting. Additionally, this approach offers insight into what constitutes an optimal model for a given hardware setting.

The main contribution of this work is the creation of a generalizable NAS framework that offers a variety of search algorithms, performance predictor solutions in a variety of objective optimization settings across several modalities. Additionally, we demonstrate how pairing evolutionary algorithms in an iterative fashion with lightly trained performance predictors can yield an accelerated and less costly exploration of a DNN architectural design space across the modalities of machine translation, recommendation, and image classification.



\section{Related Work}

\label{sec:related_work}
The computational overhead of evaluating DNN architectures during NAS can be very costly due to the training and validation cycles. To address the training overhead, novel weight-sharing approaches known as one-shot or super-networks \citep{darts2018, bender18a} have offered a way to mitigate the training overhead by reducing training times from thousands to a few GPU days \citep{elsken2019neural}. These approaches train a task-specific super-network architecture with a weight-sharing mechanism that allows the sub-networks to be treated as unique individual architectures. This enables sub-network model extraction and validation without a separate training cycle. However, the validation component still comes with a high overhead since there are many possible sub-networks which may be found from large super-networks (e.g., search space size of $\sim10^{19}$) and the validation step itself comes with a computational cost, especially for larger datasets such as ImageNet \citep{imagenet}.  One popular way to mitigate the validation cost in one-shot networks is to train predictors for objectives such as inference time (a.k.a. latency) and accuracy from a training set with thousands of sampled architectures \citep{cai2019once}. 

Tangentially, some approaches iterate training and search in order to fine-tine the super-network during training \citep{Lu_2021_NAT}. However, the fine-tuning of these approaches is influenced by the hardware platform used during search which could require training to be redone if the resulting super-network were deployed on a different platform (e.g., trained on GPU, deployed onto Raspberry Pi). To further address validation costs, novel approaches in NAS without training \citep{mellor2021neural} and meta-learning \citep{lee2021help} offer promising solutions but are designed for a single-objective settings whereas in this work we are interested in the multi-objective setting for obtaining hardware-performance trade-offs. 

In this work, we focus on demonstrating evolutionary algorithm (EA) and sequential model-based optimization (SMBO) approaches that are known to pair well with weight-sharing-based search spaces. Genetic algorithms, a subset of EA, have been broadly applied for image classification NAS problems in both single-objective implementations \citep{guo2020single} and in multi-objective approaches such as NSGA-Net \citep{lu2019nsganet}. We note that reinforcement learning (RL) and gradient optimization have found success in the NAS field as well \citep{ren2021comprehensive}.

\begin{flushleft}
\textbf{Limitations} From a framework perspective, we focus on accelerating the post-training sub-network search process, not the optional fine-tuning stage. After promising DNN architectures are discovered with NAS, a user may achieve state-of-the-art performance for a particular performance bound (e.g., top-1 accuracy for a specific latency or MACs range) by finding the right combination of fine-tuning tactics or by completely re-training the sub-network from scratch \citep{wu2021stronger}. Additionally, we do not evaluate the search performance of RL-based NAS and note \citet{efficientnas} found comparable performance to EAs.
\end{flushleft}

\begin{flushleft}
\textbf{Broader Impact} We do not anticipate that our work will have negative societal impacts. Our work leverages the one-shot weight sharing NAS paradigm which inherently provides massive savings in computation resources resulting in lower CO\textsubscript{2} emissions \citep{cai2019once}. Moreover the computational cost (i.e., energy consumption) is further reduced by our approach to accelerate the architecture search process although it remains non-trivial. In terms of data, the datasets in this work such are openly available and have been widely used in previous research (Table \ref{tab:super_network_summary}).
\end{flushleft}

\section{Search Methodology}

\subsection{Framework Description}
\label{ssec:framework_description}

\begin{figure}[t]
    \centering
    \includegraphics[width=0.95\linewidth]{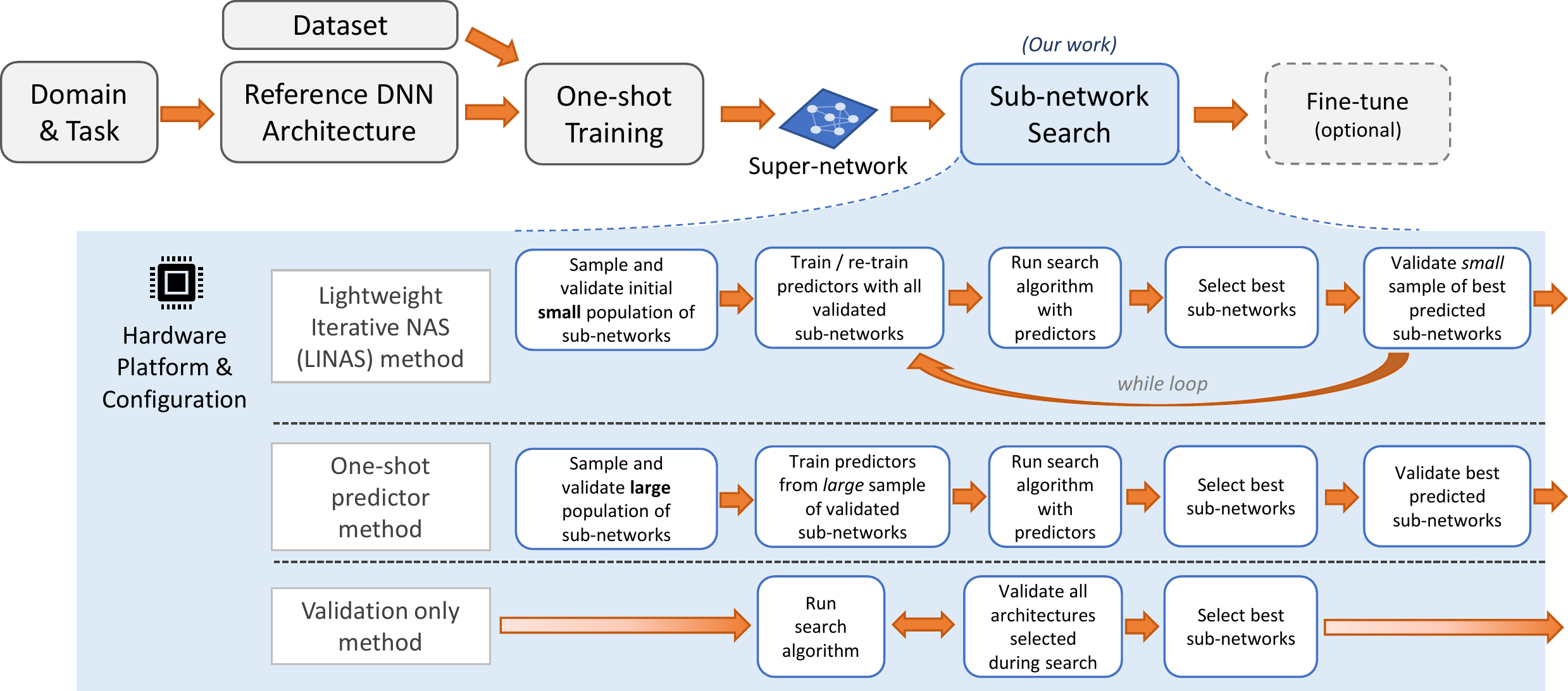}
    \caption{Generalizable framework for accelerating super-network type neural architecture search.}
    \label{fig:system_flow}
\end{figure}

Given the growing popularity of super-network DNN architectures across a plethora of machine learning problem domains, we describe a flexible hardware-aware super-network search framework in Figure \ref{fig:system_flow}. For an arbitrary super-network reference architecture, modality, and task, our system flow automates the architecture search process and discovers sub-networks that are optimal for a set of one or more performance objectives (e.g., accuracy, latency, MACs, etc.). The primary goal of this framework is to reduce the number of validation measurements (not predictions) that are required to find optimal DNN architectures given a set of performance objectives and hardware platform.

The simplest \emph{validation only} search method performs a validation measurement for sub-networks identified by the search algorithm and hence comes with a high computational cost. Even in the weight-sharing super-network NAS context, a validation measurement still requires a non-trivial amount of time and computational resources. The framework also offers a \emph{one-shot} predictor approach to reduce the validation cost overhead as described in the related work. In addition to these methods, we propose a lightweight iterative NAS (LINAS) method described in Section \ref{ssec:linas} that builds on the idea that lightly trained predictors can yield useful information for ranking sub-network configurations. The LINAS method increases the probability that optimal architectures will be identified in early stages of the search and avoids the upfront validation cost of the one-shot predictor approach. Our framework is built on top of the pymoo\footnote{https://pymoo.org} \citep{pymoo} and Optuna\footnote{https://optuna.org} \citep{optuna_2019} optimization libraries for reproducibility and ease of future algorithmic enablement. 

\subsection{Super-Network Modalities}
\label{ssec:supernetwork_modalities}

The majority of NAS research efforts have focused on the computer vision task of image classification and only recently have other modalities, such as the rapidly growing field of language modeling or language translation, been investigated in detail (\cite{wang2020hat, feng21}). Subsequently, understanding how NAS approaches generalize across modalities has not been studied in depth. In the study of our framework, our experiments encompass the modalities of image classification, machine translation, and recommendation as shown in Table \ref{tab:super_network_summary}.

For the modality of image classification, we leverage two super-networks derived from MobileNetV3 \citep{howard2019searching}  and ResNet50 \citep{resnet2016} which are described in Once-for-all (OFA) \citep{cai2019once}. OFA employs a progressive shrinking method during super-network training resulting in \emph{elastic} design parameters that can represent the full architectural search space. For additional variety in this domain, we use recent work by \citet{bootstrapNAS}, called BootstrapNAS, and discuss the sub-network search process for the quantized INT8 space in the Appendix.

Super-network approaches have recently been applied in the domain of Natural Language Processing (NLP). Hardware-aware Transformers (HAT) \citep{wang2020hat} achieve this goal by extending the network elasticity type of weight sharing approach to this domain. In HAT, the authors introduce \textit{arbitrary encoder-decoder attention}, to break the information bottleneck between the encoder and decoder layers in Transformers \citep{vaswani2017attention}. Additionally, they propose heterogenous transformer layers to allow for different layers to have different parameters.  

Neural Collaborative Filtering (NCF) \citep{he2017neural}, a popular method for recommendation problems, combines the benefits of traditional matrix factorization and fully connected neural networks. We adapt this model architecture into an elastic training framework similar to HAT wherein each embedding layer and dense layer is fully elastic.

\begin{table*}[tb]
	\caption{Summary of super-networks and associated design search spaces used in this work. Additional details for each super-network are provided in Appendix \ref{sec:appendix_supernets}.}
	\centering
	\small
	\begin{tabular}{cccccc}
	\hline \hline
	Super-Network & Task & Dataset & \begin{tabular}[c]{@{}c@{}}Number \\ Format\end{tabular} & Objectives & \begin{tabular}[c]{@{}c@{}}Search Space Size \\ (unique DNNs)\end{tabular} \\
	\hline \hline
	MobileNetV3 & \begin{tabular}[c]{@{}c@{}}Image \\ Classification\end{tabular} & ImageNet & FP32 & \begin{tabular}[c]{@{}c@{}}Top-1 Accuracy, \\ Latency\end{tabular} & $\sim 10^{19}$ \\
	\hline
	ResNet50 & \begin{tabular}[c]{@{}c@{}}Image \\ Classification\end{tabular} & ImageNet & \begin{tabular}[c]{@{}c@{}}FP32\end{tabular} & \begin{tabular}[c]{@{}c@{}}Top-1 Accuracy, \\ Latency\end{tabular} & $\sim10^{13}$ \\
	\hline
	Transformer & \begin{tabular}[c]{@{}c@{}}Machine \\ Translation\end{tabular} & \begin{tabular}[c]{@{}c@{}}WMT 2014 \\ En-De\end{tabular} & FP32 & \begin{tabular}[c]{@{}c@{}}BLEU Score, \\ Latency\end{tabular} & $\sim10^{15}$ \\
	\hline
	NCF & Recommendation & Pinterest-20 & FP32 & \begin{tabular}[c]{@{}c@{}}HR@10, \\ Latency\end{tabular} & $\sim10^{7}$ \\
	\hline \hline
	\end{tabular}
	\label{tab:super_network_summary}
\end{table*}

\subsection{Search Space Encoding}

\begin{figure}[htb]
  \centering
  \begin{subfigure}[t]{0.48\linewidth}
    \centering
    \includegraphics[width=1.0\linewidth]{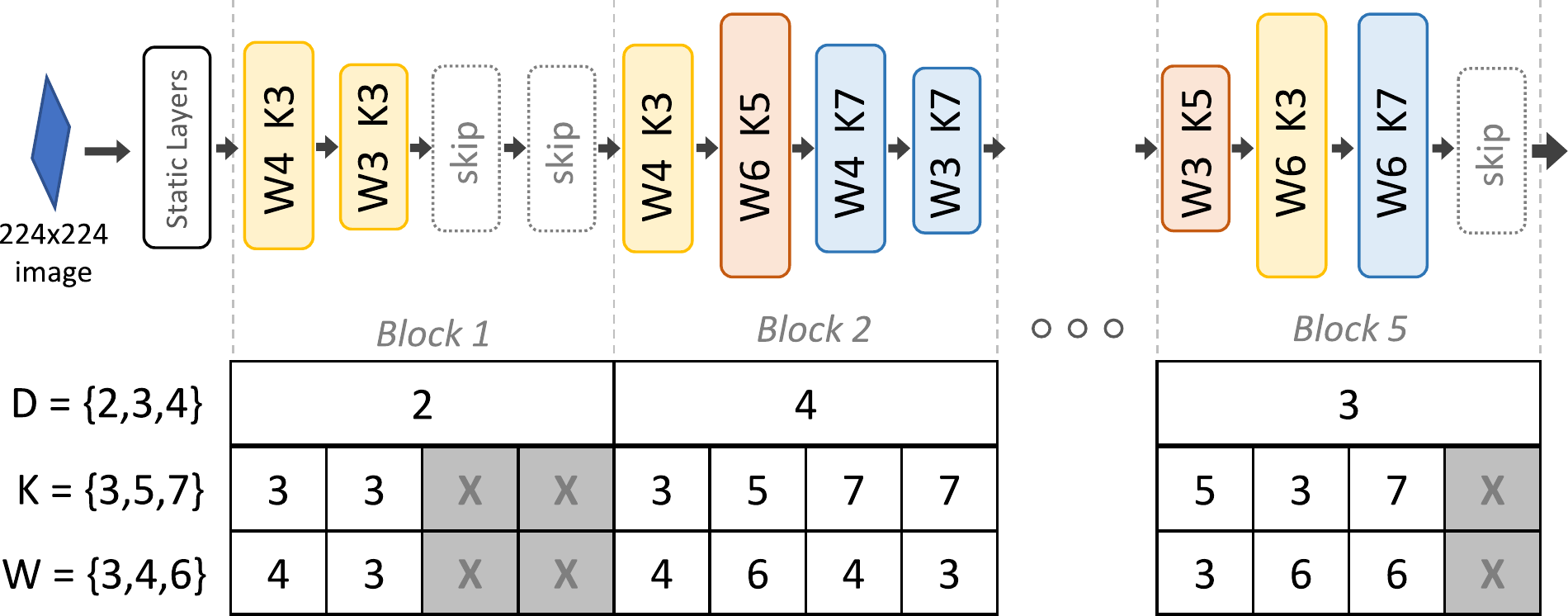}
    \caption{MobileNetV3 design variable encoding.}
    \label{sfig:mbv3_encoding}
  \end{subfigure}
  \hfill
  \begin{subfigure}[t]{0.48\linewidth}
    \centering
    \includegraphics[width=1.0\linewidth]{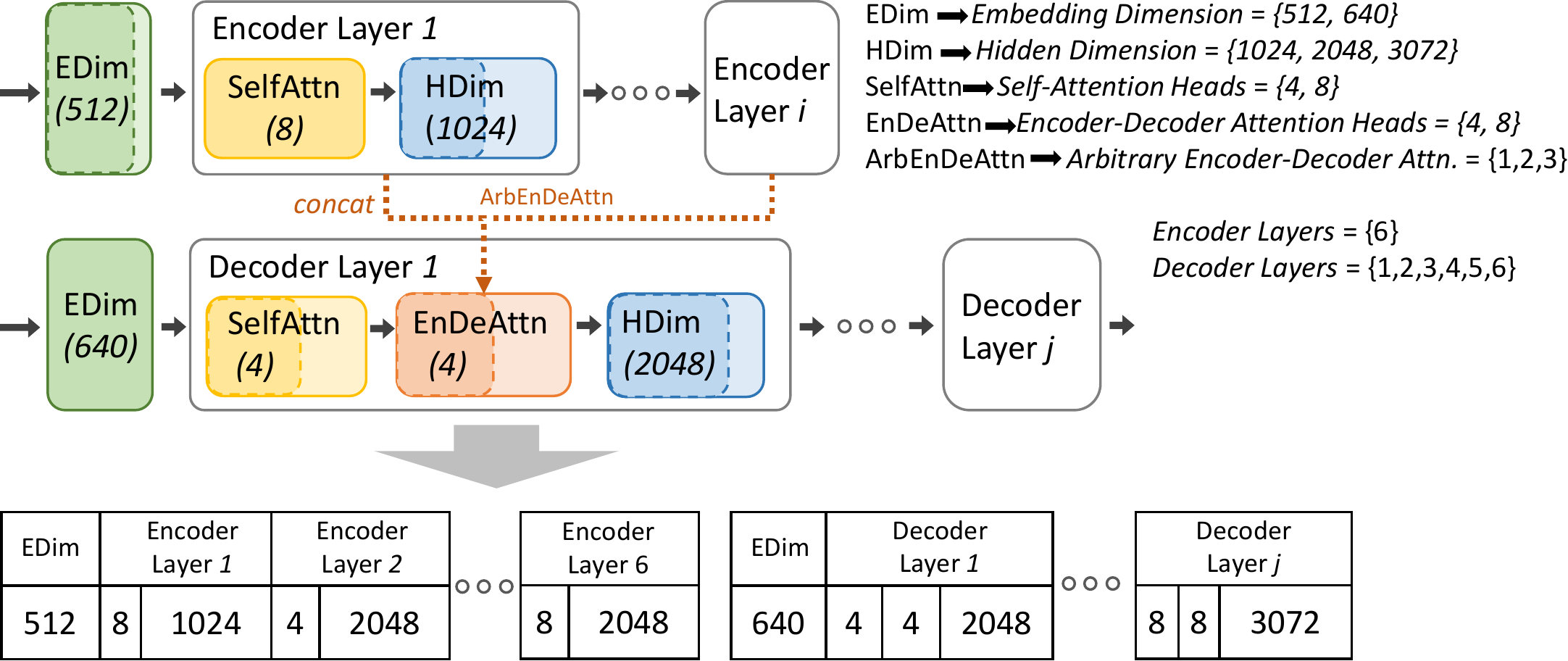}
    \caption{Transformer design variable encoding.}
    \label{sfig:transformer_encoding}
  \end{subfigure}
  \caption{Super-network encoding strategies for MobileNetV3 and Transformer each having 45 and 40 design variables respectively.}
  \label{fig:encoding}
\end{figure}

A key consideration of the super-network NAS process is encoding a representation of the architectural design variables in a way that is useful for the search algorithms. For illustration, we summarize our encoding strategy for the MobileNetV3 and Transformer super-networks in Figure \ref{fig:encoding}. By mapping each super-network architecture design variable to several integer options, this search space encoding offers a compatible interface for the evolutionary operators (e.g., mutation, crossover, etc.). 

\subsection{Predictors}
\label{ssec:predictors}

Since validation evaluations of performance objectives, such as top-1 accuracy and latency, require a large amount of time, we follow the work in \citep{cai2019once} and \citep{wang2020hat} and employ predictors. More specifically, we predict top-1 accuracy of sub-networks derived from MobileNetV3 and ResNet50 super-networks, hit ratio (HR@10) of sub-networks derived from NCF super-networks, bilingual evaluation understudy (BLEU) \citep{papinenibleu2002} score of sub-networks derived from Transformer super-networks and latency of sub-networks derived from all super-networks.

However, unlike prior work which use multi-layer perceptrons (MLPs) to perform prediction, we employ much simpler methods such as ridge regression, support vector machine regression (SVR) and stacked regression predictors. The authors of \citep{Lu_2021_NAT} and \citep{laube2022what} have found that MLPs are inferior to other methods of prediction for low training example counts. We have found that these simpler methods converge more quickly and require both fewer training examples and much less hyper-parameter optimization than MLPs. The combination of performance objective prediction via simple predictors allows us to significantly accelerate the selection of sub-networks with minimal prediction error. See Section \ref{ssec:accuracy_predictor_analysis} for an analysis of our predictor performance.

\subsection{Search Algorithms}
\label{ssec:linas}

The foundational goal of NAS is to find DNN architectures that are optimal for one or more performance objectives. In the context of weight-sharing super-networks, consider a pre-trained super-network with weights $W$, a set of sub-network architectural configurations $\Omega$ derived from the super-network and $m$ competing objectives $f_1(\omega; W), \ldots, f_m(\omega; W)$ where $\omega \in \Omega$. Each of the sub-network configurations $\omega$ is a valid set of parameters used during training of the super-network. For example, a given $\omega$ will contain values for each design parameter (e.g., depth and kernel size) used during super-network training. Our system aims to minimize a subset of objectives $S_i \subseteq \{f_1(\omega; W), \ldots, f_m(\omega; W)\}$ to discover the near-optimal sub-network $\omega_i^*$. In other words,
\begin{equation}
    \omega_i^*=\underset{\omega \in \Omega}{\text{argmin}}\big(S_i\big)
\end{equation}
An objective can be negated to transform a minimization objective into a maximization objective (e.g., accuracy is a maximization objective). During optimization, multiple architectures $\omega_i^* \in \Omega$ will be scored in the objective space allowing for the identification of a \textit{Pareto front}, as illustrated in Figure \ref{fig:hypervolume_cartoon}.

In this work we focus on examining random search, multi-objective sequential model-based optimization (SMBO), and multi-objective evolutionary algorithm (MOEA) approaches to the sub-network search problem. From a hardware-aware standpoint, we evaluate in the multi-objective (a.k.a. bi-objective) setting as we are focused on finding a highly diverse set of near-optimal architectures across the accuracy and latency trade-off (Pareto front) region. However, we note that our framework works with any number of objectives. To test a SMBO algorithm in our framework we employ the multi-objective tree-structured parzen estimator (MOTPE) as proposed by \citet{motpes}. From the Pareto-based MOEA category, the framework supports the popular NSGA-II \citep{deb2002fast} algorithm and a similar approach called AGE-MOEA \citep{agemoea}. For indicator- and decomposition-based MOEAs we support U-NSGA-II \citep{deb2006}, MOEA/D \citep{moead}, and CTAEA \citep{ctaea}. 

One of the primary goals for our framework is to reduce the number of validation measurements that are required to find optimal DNN architectures in a multi-objective search space that works well across modalities. While related work shows that using trained predictors can speed up the DNN architecture search process, there remains a substantial cost to training the predictors since the number of validated training samples can range between 1000 and 16,000 \citep{lu2020nsganetv2}. Interestingly, as shown in Figure \ref{fig:accuracy_predictor_analysis}, simple accuracy predictors can achieve acceptable mean absolute percentage error (MAPE) with far fewer training samples. We build on this insight that lightly trained predictors can offer a useful surrogate signal during search. Algorithm \ref{alg:linas} describes our generalizable Lightweight Iterative NAS (LINAS) method. We first randomly sample the architecture search space to serve as the initial validation population. For each sub-network in the validation population, we measure each objective and store the result. These results are combined with all previous validation results and are used to train the objective predictors. For each iteration, we run a multi-objective algorithm search (e.g., NSGA-II) using that iteration's trained predictors for a high number of generations (e.g., $>100$) to allow the algorithm to explore the predicted objective space sufficiently. This predictor-based search runs very quickly since no validation measurements occur. Finally, we select the most optimal population of diverse DNN architectures from the predictor-based search to add to the next validation population, which then informs the next round of predictor training. This cycle continues until the iteration count limit is met or an end-user decides a sufficient set of architectures has been discovered. We note that the LINAS approach can be applied with any single-, multi-, or many-objective EA and generalizes to work with any super-network framework. Additionally, it allows for the interchanging of tuning parameters (e.g., crossover, mutation, population), EAs, and predictor types for each iteration.

\begin{algorithm}[ht]
   \caption{Generalizable Lightweight Iterative Neural Architecture Search (LINAS)}
   \label{alg:linas}
\begin{algorithmic}
   \STATE {\bfseries Input:}
   Objectives $f_m$, super-network with weights $\mathcal{W}$ and configurations $\Omega$, predictor model for each objective $Y_{m}$, LINAS population $P$ size $n$, number of LINAS iterations $I$, evolutionary algorithm $\mathcal{E}$ with number of evaluations $J$.
   \STATE $P_{i=0} \leftarrow \{\omega_{n}\} \in \Omega$ \textcolor{gray}{// sample $n$ sub-networks for first population}\\ 
   \WHILE{$i++ < I$}
   \STATE $D_{i,m} \leftarrow f_m(P_{i} \in \Omega; \mathcal{W})$ \textcolor{gray}{// measure objectives $f_m$, store data $D_{i,m}$} \\
   \STATE $D_{all,m} \leftarrow D_{all,m} \cup D_{i,m}$
   \STATE $Y_{m,pred} \leftarrow Y_{m,train}(D_{all,m})$ \textcolor{gray}{// train predictors for each objective} 
   \WHILE{$j++ < J$} 
   \STATE $P_{\mathcal{E}_{j}} \leftarrow \mathcal{E}(Y_{m,pred}, j)$ \textcolor{gray}{// run ${\mathcal{E}}$ for $J$ evaluations}
   \ENDWHILE
   \STATE $P_{i} \leftarrow P_{\mathcal{E},best\_unique} \in P_{\mathcal{E}_{J}}$ \textcolor{gray}{// retrieve optimal and unique population of sub-networks}
   \ENDWHILE
   \STATE {\bfseries Output:} All validated sub-networks configurations $P_{I}$, predictor search results $P_{\mathcal{E}_{I,J}}$, and validation data $D_{all,m}$. 
   
\end{algorithmic}
\end{algorithm}

\section{Experiments \& Results}

In the hardware-aware NAS context, latency is a highly important metric since it directly relates to the real-time performance of a hardware system. Often MACs, FLOPs, or model parameter counts are used as an approximation of latency but do not guarantee correlation (see Appendix \ref{sec:appendix_macs}). We use the term "hardware-aware" to emphasize the focus on using latency as one of our main objectives but do not use any hardware architectural information to inform the search process. In this work we experiment on CPU, GPU, and mobile device platforms for evaluating our framework and we describe the transferability of NAS results between CPU and GPU platforms in Appendix \ref{sec:hardware_platform_transfer}. Since our experiments measure latency values from different manufacturers and there are possible proprietary issues in sharing what could be perceived as official benchmark data, we normalize latency results to be within $[0, 1]$. More specifically, the normalized latency $\hat{l}$ is given by $\hat{l}=\frac{l-l_{min}}{l_{max}} \in [0,1]$ where $l$ is the unnormalized latency, $l_{min}$ is the minimum unnormalized latency and $l_{max}$ is the maximum unnormalized latency. Using normalized latency does not change the underlying search results we are demonstrating. For comparative latency performance metrics related to our test platforms, we point the reader to the MLCommons\footnote{https://mlcommons.org} benchmark suite.

\subsection{Predictors}
\label{ssec:accuracy_predictor_analysis}

As described in Section \ref{ssec:predictors}, our work makes extensive use of predictors to accelerate the selection of sub-networks, particularly when applying LINAS. Predictors are necessary since performing actual measurements of performance objectives such as accuracy or latency would be prohibitively slow. In light of their importance, a better understanding of their performance is needed.

\begin{figure}[htb]
    \centering
    \includegraphics[width=1.0\linewidth]{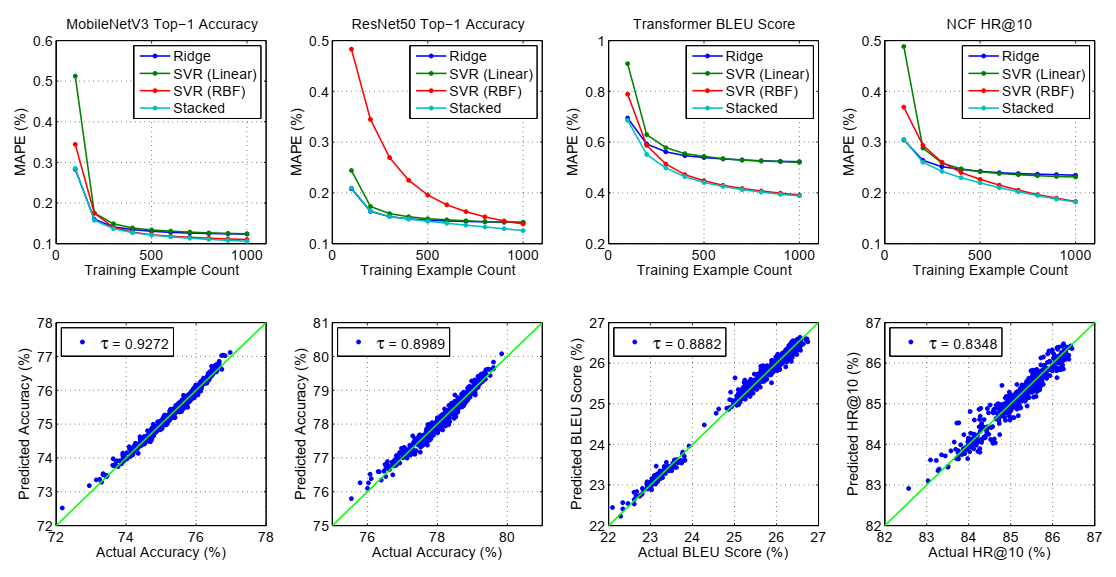}
    \caption{MAPE of predictors performing top-1 accuracy, BLEU score and HR@10 prediction versus the number of training examples for sub-networks derived from the super-networks shown in Table \ref{tab:super_network_summary} (top row). Correlation and Kendall $\tau$ coefficient between actual and predicted values after training the stacked predictor with 1000 examples (bottom row). The ideal correlation is shown by the green line.}
    \label{fig:accuracy_predictor_analysis}
\end{figure}

The analysis of predictors is performed over a number of different trials to account for variance in the results. In each trial, the data set for each predictor is first split into train and test sets. Subsets of the train data set within the range of 100 to 1000 examples are used to train the predictor. For a given trial, the \textit{same} test set with 500 examples is used to compute the prediction mean absolute percentage error (MAPE). This process is repeated for a total of 100 trials and the results averaged to compute the MAPE shown in Figure \ref{fig:accuracy_predictor_analysis}.

The top row of Figure \ref{fig:accuracy_predictor_analysis} shows the MAPE of different predictors for each super-network type in Table \ref{tab:super_network_summary}. The \textit{stacked} predictor is a combination of ridge and SVR (RBF) regressors which "stacks" the predictions from each of these two regressors and uses them as the input to a final ridge regressor. The bottom row shows the correlation between actual and predicted values after training the stacked predictor with 1000 examples. Note that the Kendall rank correlation coefficient $\tau$ is also shown for each case. In all cases, these simple predictors provide small error (maximum MAPE of 0.91\%) and high correlation (minimum $\tau$ of 0.8348) with actual values. Results for latency prediction are shown in Appendix \ref{sec:latency_prediction}.

\subsection{Search Results}
\label{ssec:linas_results}

\begin{figure}[htb]
    \centering
    \includegraphics[width=0.3\linewidth]{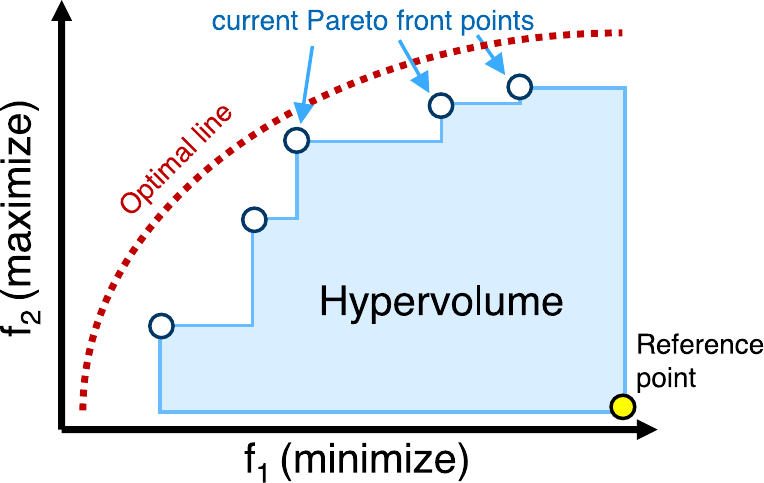}
    \caption{Illustration of the hypervolume metric in a bi-objective space.}
    \label{fig:hypervolume_cartoon}
\end{figure}

The main purpose of our framework and proposed LINAS approach is to reduce the total number of validation measurements required to find optimal DNN architectures in the multi-objective space for any modality or domain-specific task. In other words, the goal of LINAS is to maximize the hypervolume while minimizing the number of evaluations during NAS. Specifically, we want to efficiently discover architectures with optimal trade-offs in high top-1 accuracy/BLEU/HR@10 and low latency. Since we are interested in evaluating the performance of various search algorithms in the multi-objective setting, we use the hypervolume indicator \citep{zitzler1999multiobjective} as shown in Figure \ref{fig:hypervolume_cartoon}. When measuring two objectives, the hypervolume term represents the dominated \textit{area} of the Pareto front. In our experiments we use the term evaluation to refer to an actual validation measurement, not a predicted measurement. 

\begin{figure}[htb]
  \centering
  \begin{subfigure}[t]{0.32\linewidth}
    \centering
    \includegraphics[width=1.0\linewidth]{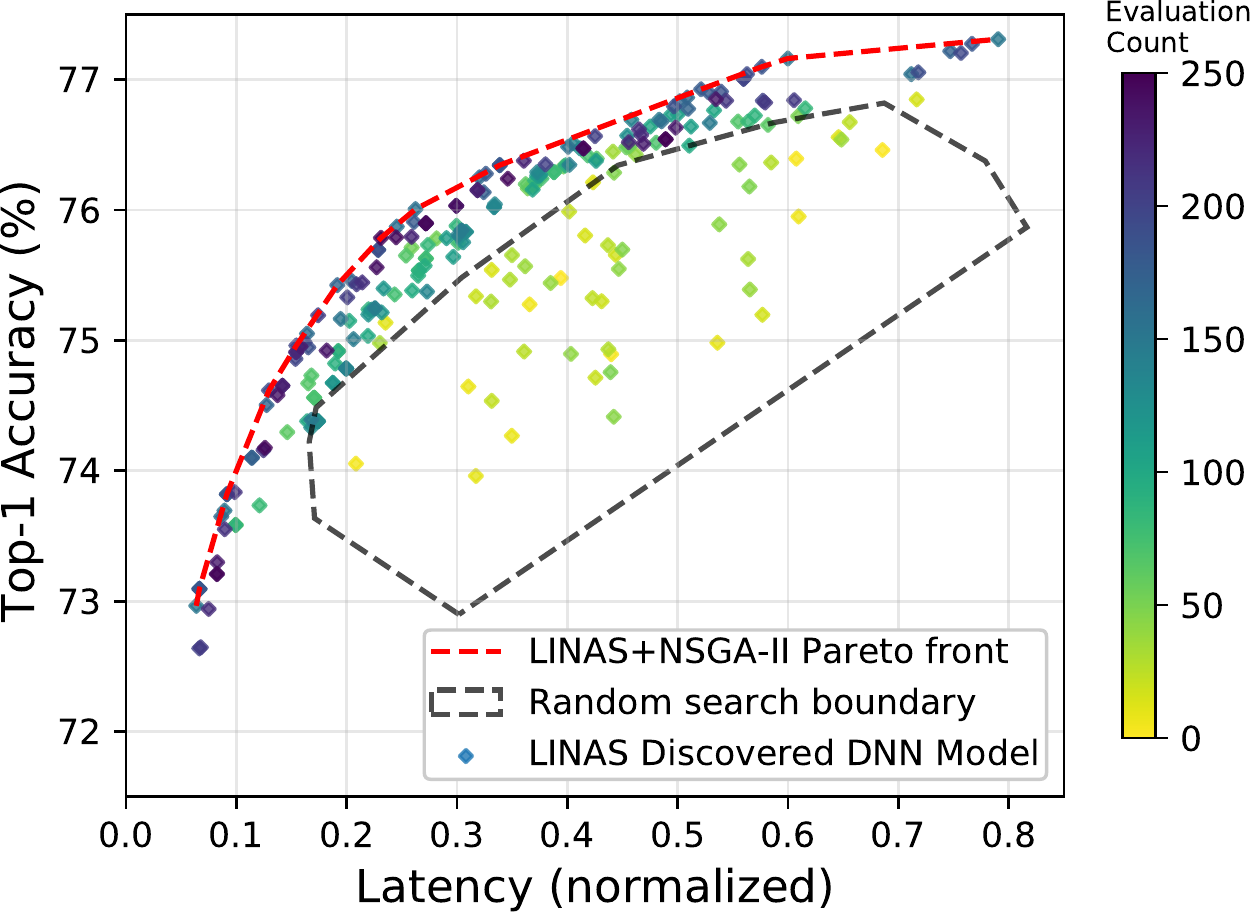}
    \caption{LINAS Search.}
    \label{subfigure_left}
  \end{subfigure}
  \hfill
  \begin{subfigure}[t]{0.32\linewidth}
    \centering
    \includegraphics[width=1.0\linewidth]{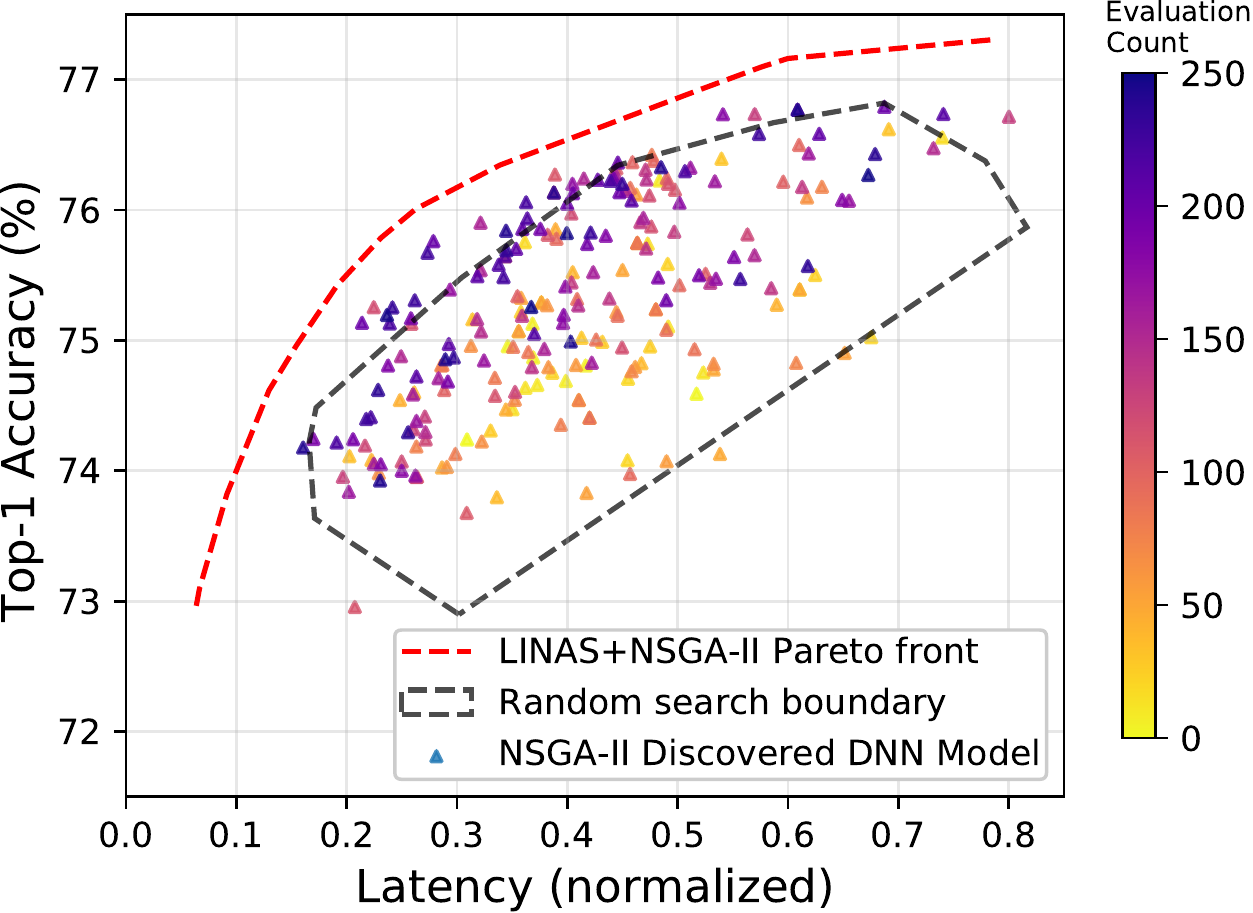}
    \caption{NSGA-II Search.}
    \label{subfigure_mid}
  \end{subfigure}
  \hfill
  \begin{subfigure}[t]{0.32\linewidth}
    \centering
    \includegraphics[width=1.0\linewidth]{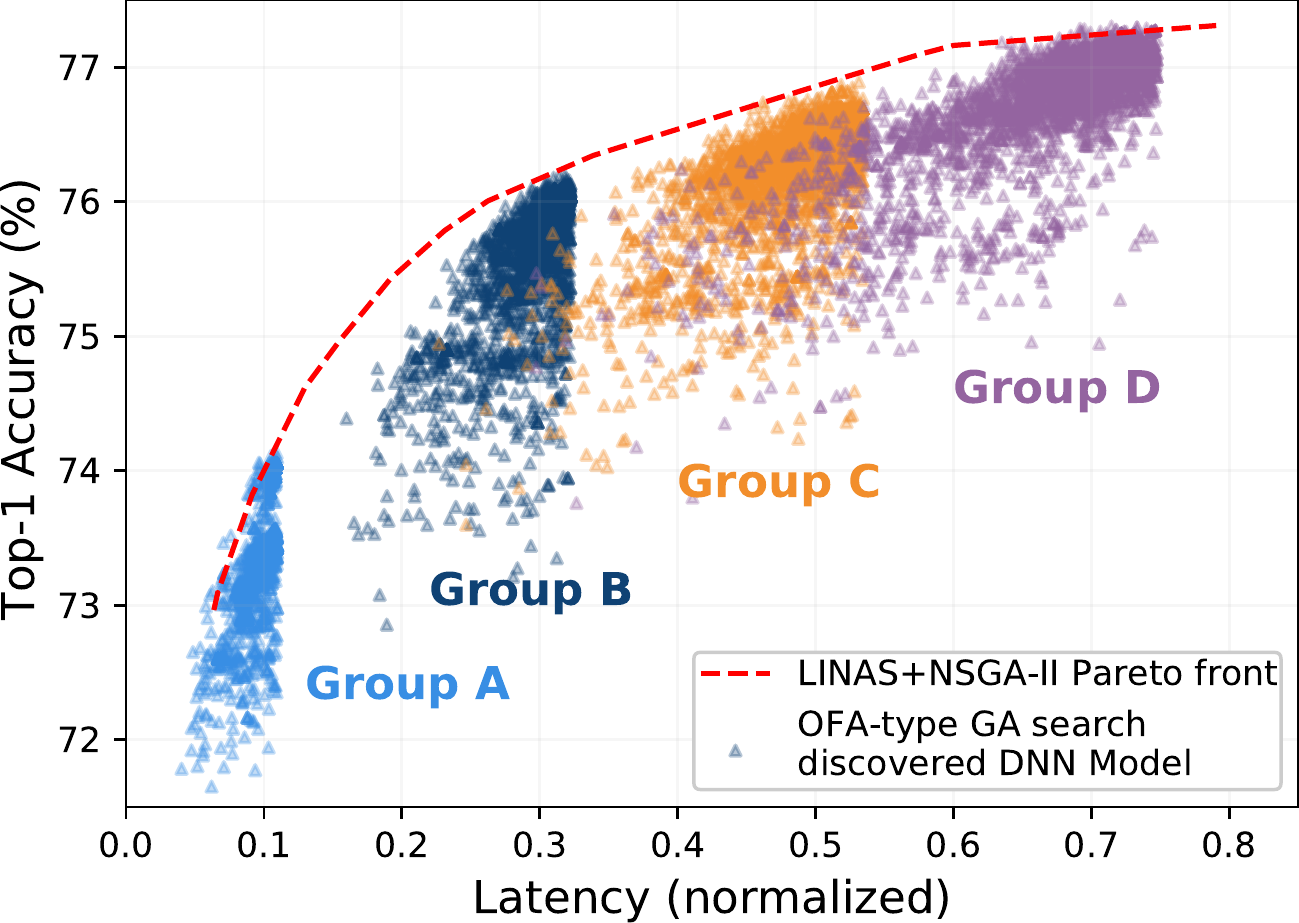}
    \caption{GA search used in OFA paper.}
    \label{subfigure_ofa}
  \end{subfigure}
  \caption{Search results in the MobileNetV3 search space (Titan-V GPU, batch size = 128) comparing \subref{subfigure_left}: LINAS, and \subref{subfigure_mid}: NSGA-II approaches (algorithm settings in Table \ref{tab:search_settings}). Sub-figure \subref{subfigure_ofa} illustrates the GA approach used in the OFA paper that uses predictors trained from 1000 evaluations where we run four different latency constrained searches.}
  \label{fig:mbnv3_scatter_plots}
\end{figure}

We start our experimental analysis using the MobileNetV3 super-network since it offers the largest search space size (e.g., $10^{19}$). Figures \ref{fig:mbnv3_scatter_plots}a and \ref{fig:mbnv3_scatter_plots}b illustrate the differences in how LINAS (with NSGA-II for the internal predictor loop), random search, and NSGA-II progress in the multi-objective search space. For the same evaluation count of 250, while NSGA-II begins progressing towards an optimal trade-off region, the LINAS results show how the exploration can be accelerated. Moreover, one can see how both approaches perform better than random search. 

Since MobileNetV3 is an OFA super-network, we run the genetic algorithm (GA) search as used in the OFA paper and show the results in Figure \ref{fig:mbnv3_scatter_plots}c for comparison. This approach follows the one-shot predictor method as shown in Figure \ref{fig:system_flow} where predictors for the objectives are trained with 1000 samples before the search starts in this setup. The search then runs a large amount of predictor-based evaluations in the latency range of interest. The intent of the OFA GA search algorithm is to maximize the accuracy for a particular latency constraint. In the multi-objective setting this has a few limitations such as needing prior knowledge of the latency space and requiring a user to manually define separate search groups across the known latency range that are unique to each hardware platform. In our Figure \ref{fig:mbnv3_scatter_plots}c example we define four search groups (each with unique latency constraints) and note that LINAS only requires 250 evaluations to find a more diverse Pareto front compared to the GA search from the OFA paper which uses 1000 evaluations to build predictors. A key takeaway is that LINAS can be used to extend the search capabilities of any super-network or weight-sharing NAS framework in the multi-objective setting. 

\begin{figure}[htb]
  \centering

    \begin{subfigure}[t]{0.33\linewidth}
    \centering
    \includegraphics[width=1.0\linewidth]{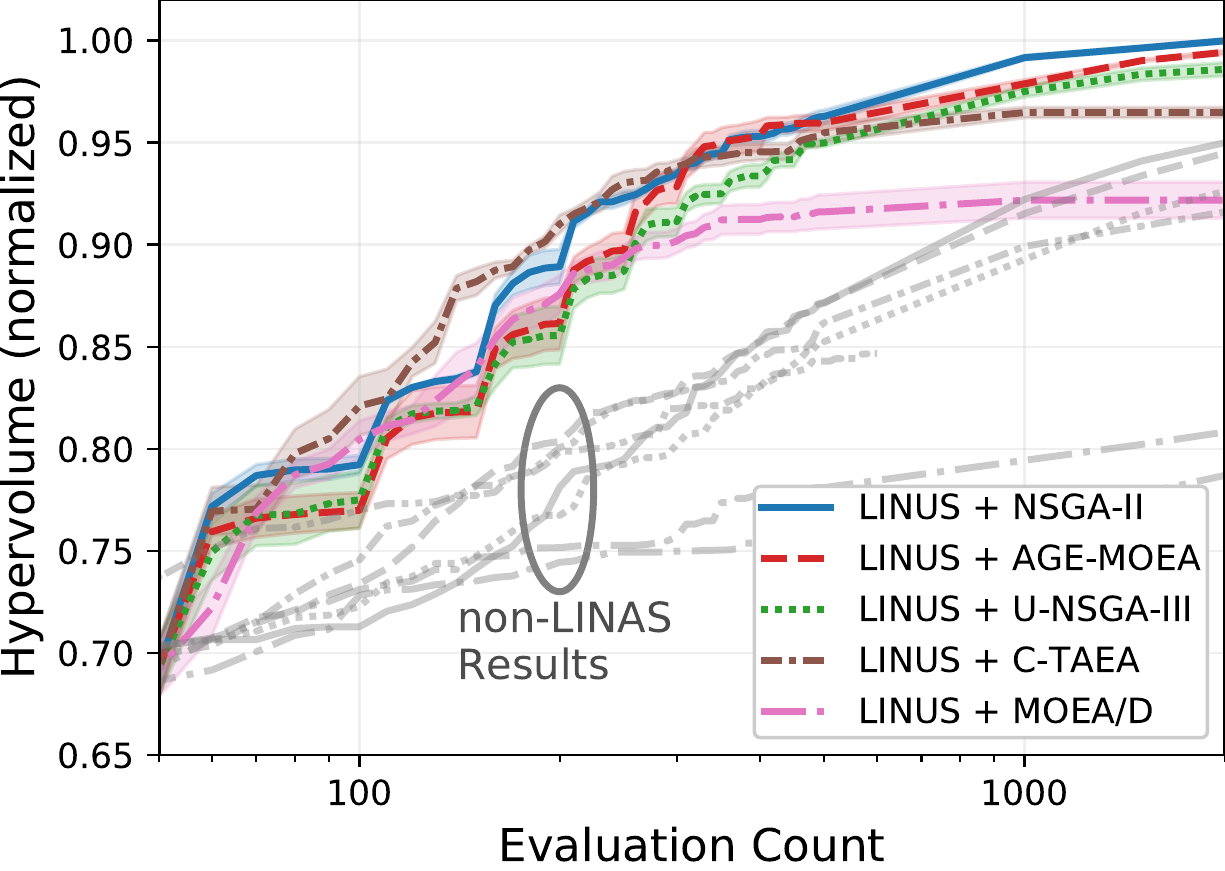}
    \caption{LINAS comparison using various EAs for the predictor loop.}
    \label{sfig:search_algorithm_comparison_linas}
  \end{subfigure}
  \hfill
  \begin{subfigure}[t]{0.33\linewidth}
    \centering
    \includegraphics[width=1.0\linewidth]{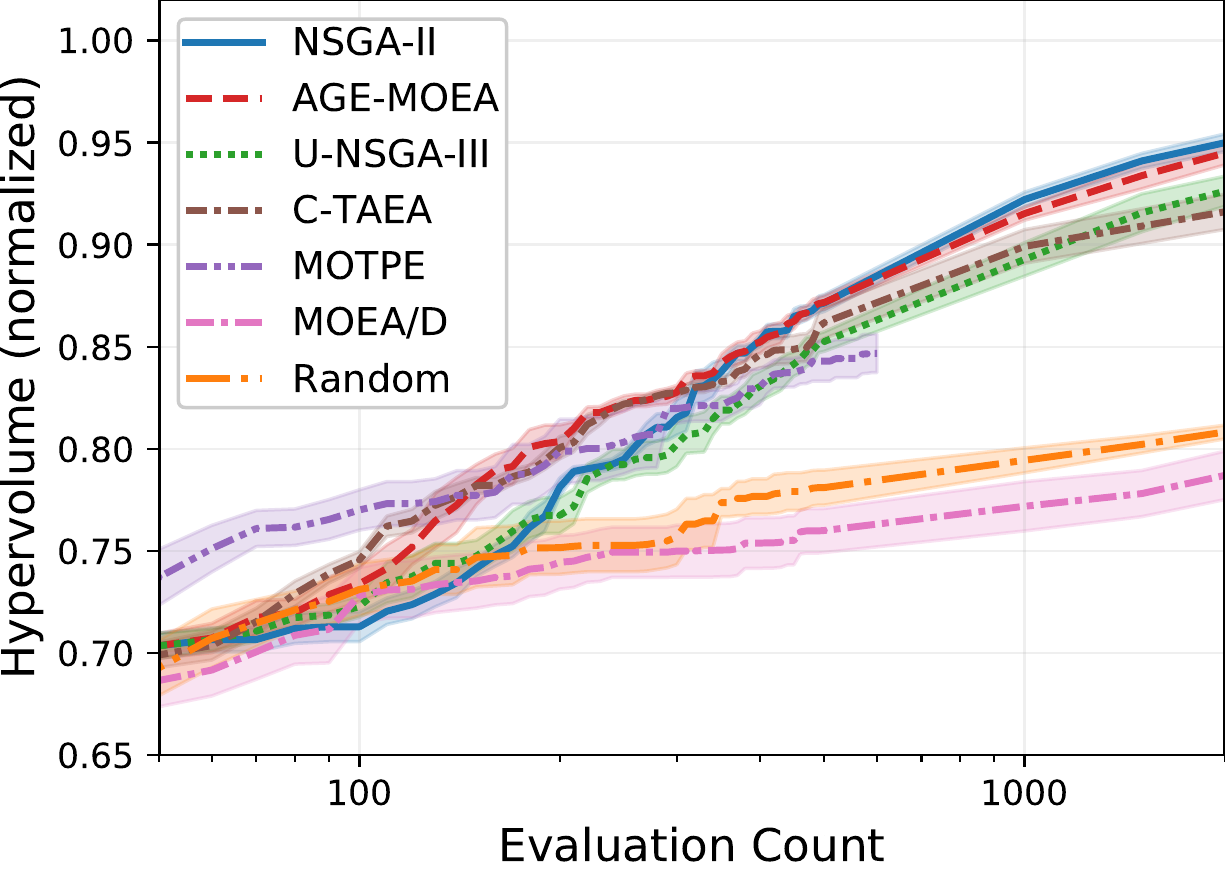}
    \caption{Search algorithm comparison without using LINAS. }
    \label{sfig:search_algorithm_comparison}
  \end{subfigure}
  \hfill
  \begin{subfigure}[t]{0.27\linewidth}
    \centering
    \includegraphics[width=1.0\linewidth]{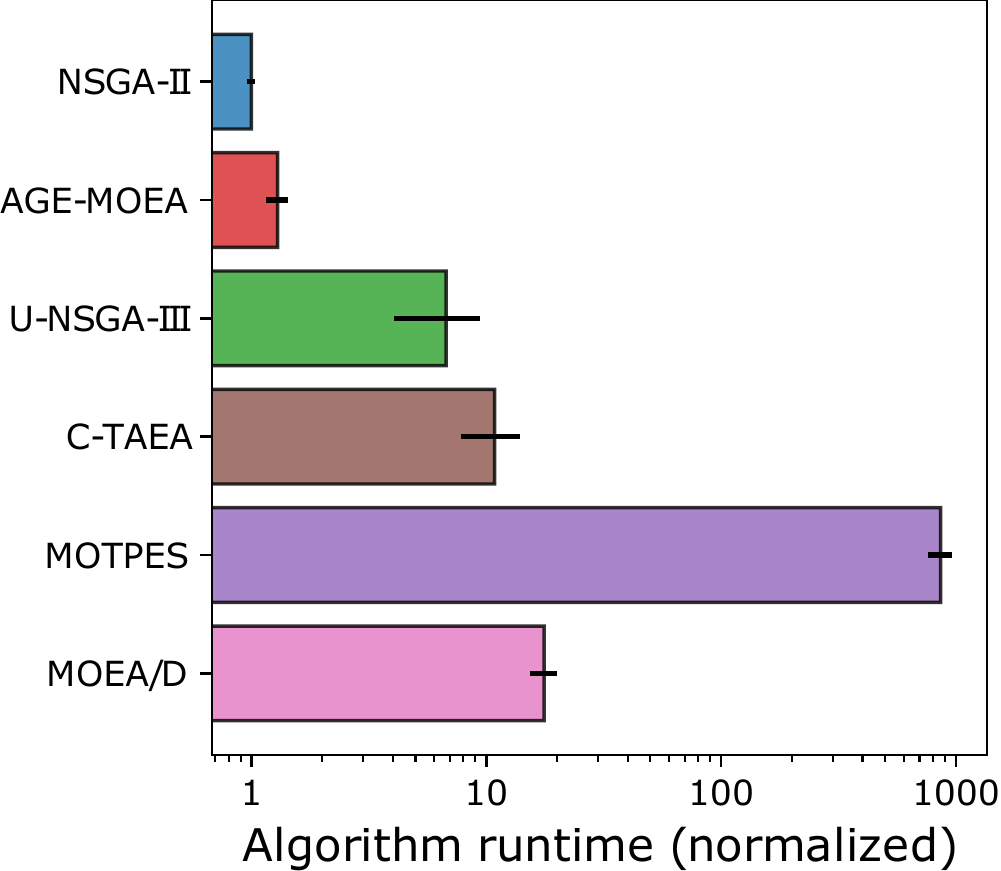}
    \caption{Average algorithm runtime.}
    \label{sfig:average_algorithm_runtime}
  \end{subfigure}
  \caption{Comparison of search algorithms in the MobileNetV3 design space for hypervolume (top-1 accuracy and latency) versus evaluation count. Shaded regions show the standard error for 5 trials with different random seeds. Search parameter settings in Appendix \ref{sec:search_algorithm_settings}.}
  \label{fig:ea_tournament}
\end{figure}

LINAS offers a great deal of flexibility in terms predictor and algorithm options for the internal loop. Figure \ref{sfig:search_algorithm_comparison_linas} shows a LINAS specific ablation study using the various EA algorithms for the internal predictor loop including the performance of various algorithms without LINAS. We find that Pareto based MOEAs such as NSGA-II and AGE-MOEA and the indicator based U-NSGA-III perform well for this task. MOTPE by itself finds good sub-networks in the very early stage of the search process but suffers from very high run-times for evaluation counts above 500. This limits the ability of MOTPE to efficiently be used in the LINAS internal predictor loop since it will not approach the near-optimal Pareto region in the predictor space. 

\begin{figure}[htb]
  \centering
  \begin{subfigure}[t]{0.32\linewidth}
    \centering
    \includegraphics[width=1.0\linewidth]{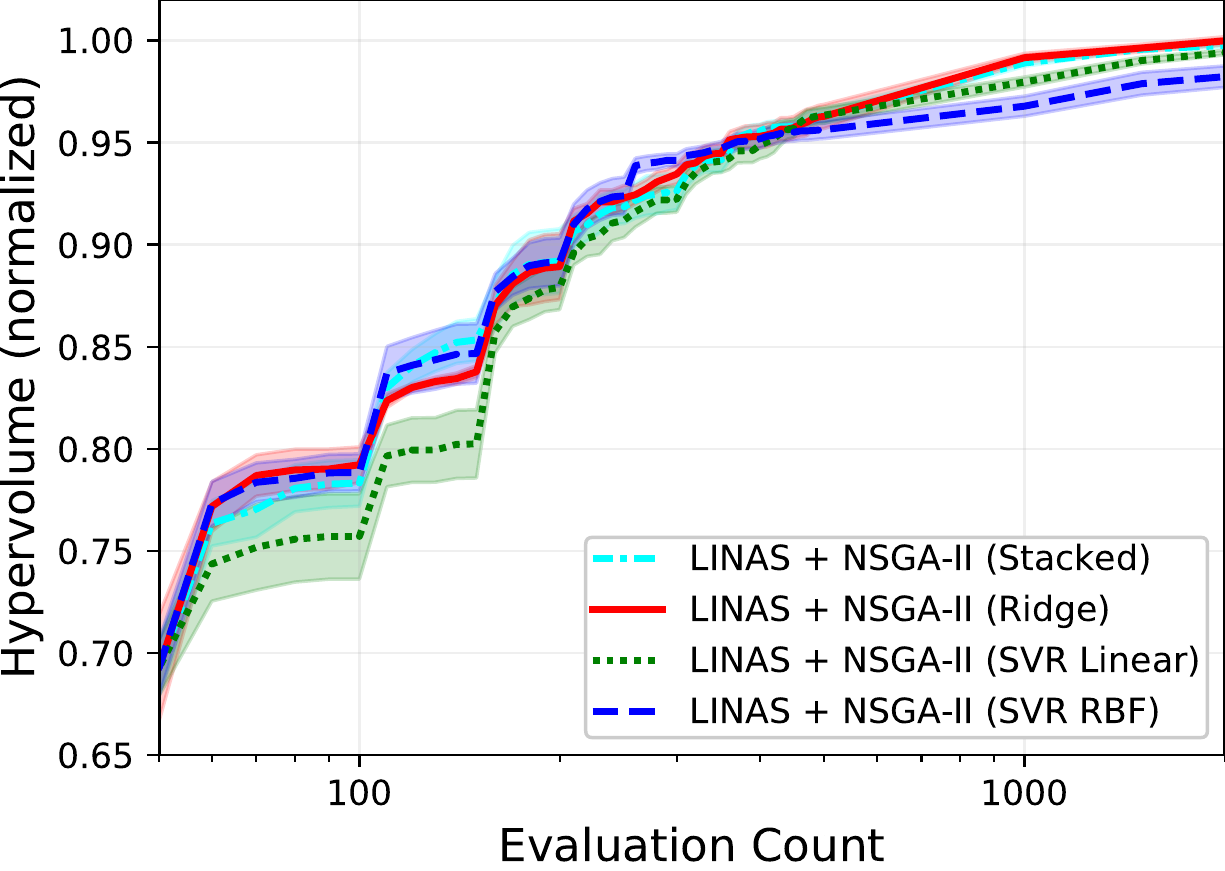}
    \caption{Predictor ablation study.}
    \label{sfig:ablation_predictors}
  \end{subfigure}
  \hfill
  \begin{subfigure}[t]{0.32\linewidth}
    \centering
    \includegraphics[width=1.0\linewidth]{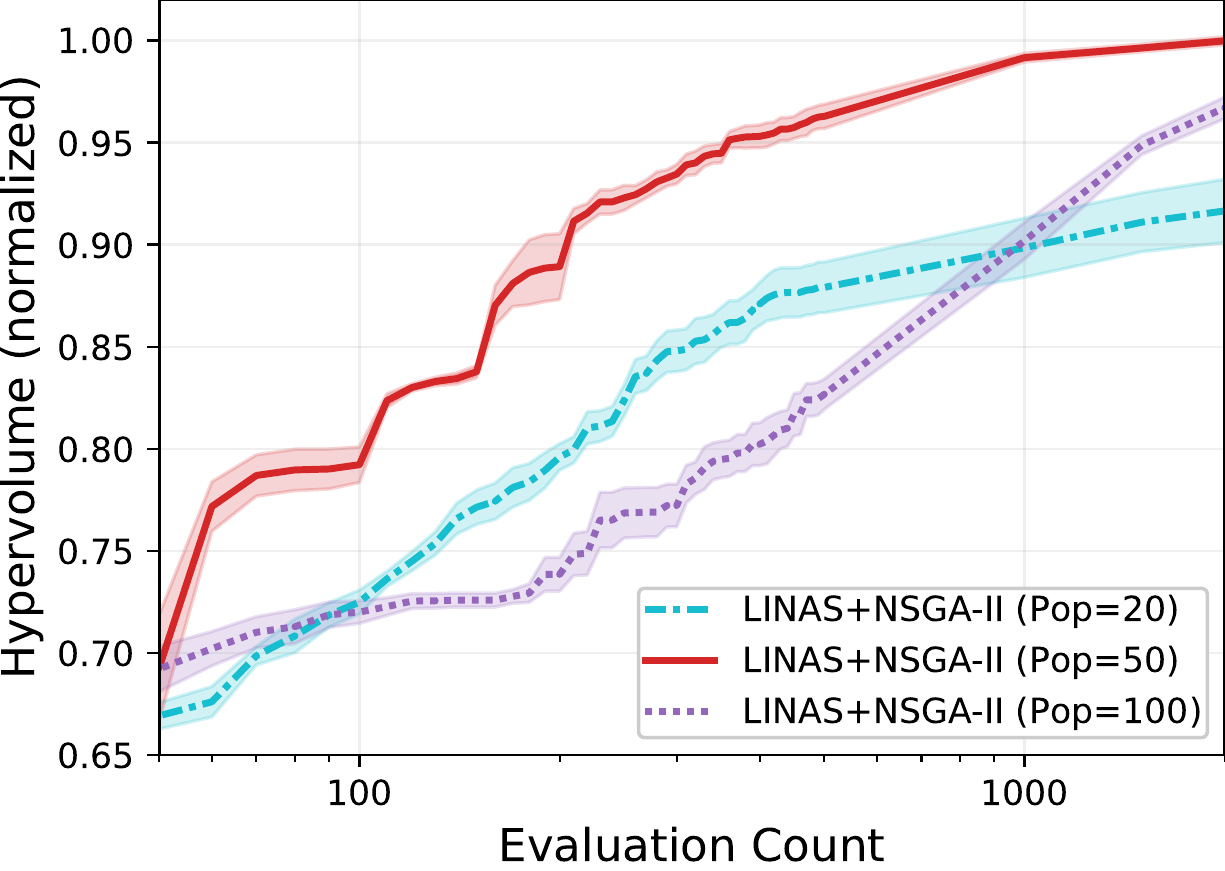}
    \caption{Population size study.}
    \label{sfig:ablation_population}
  \end{subfigure}
  \hfill
  \begin{subfigure}[t]{0.32\linewidth}
    \centering
    \includegraphics[width=1.0\linewidth]{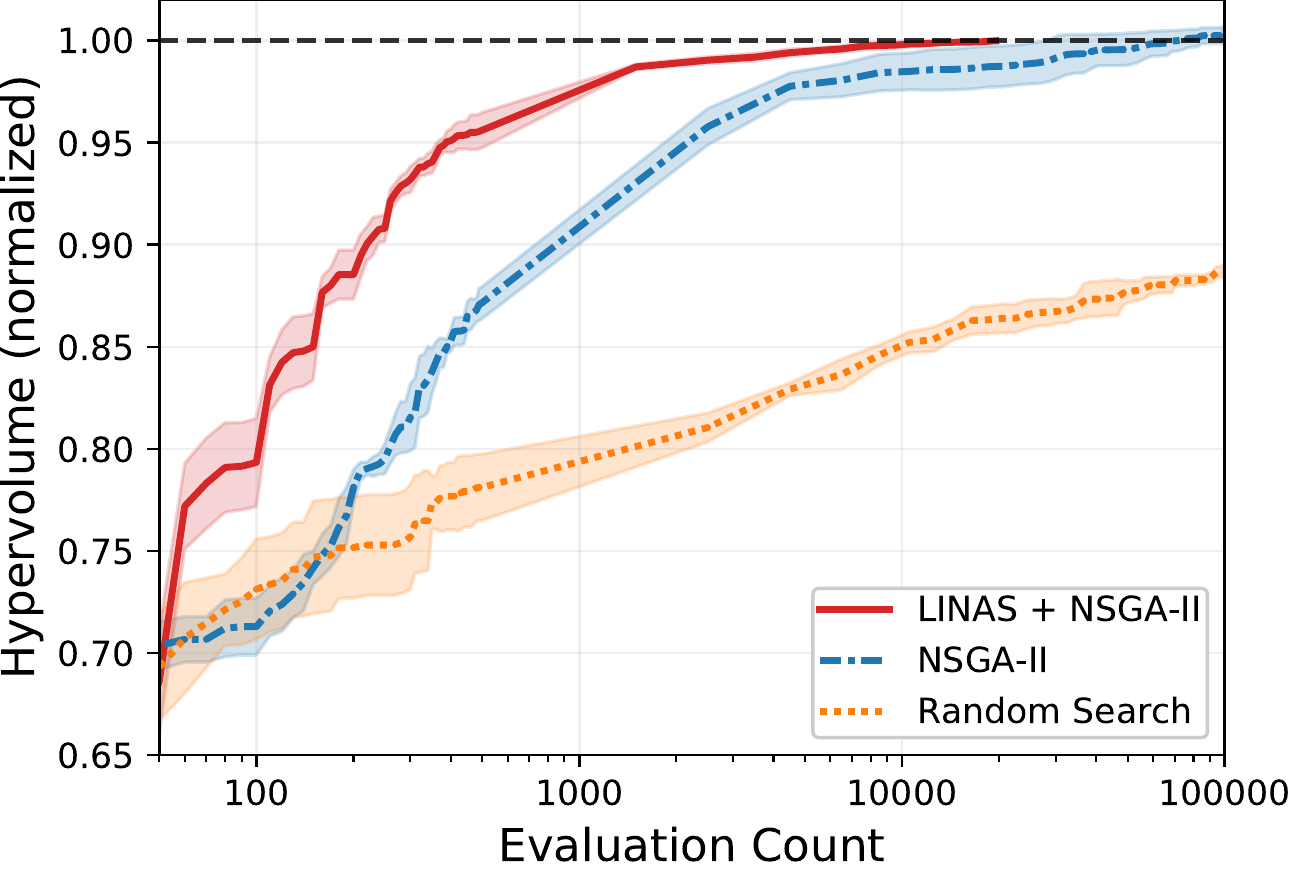}
    \caption{Extended run example}
    \label{sfig:long_search}
  \end{subfigure}
  \caption{Hypervolume ablations studies on MobileNetV3 (Titan-V GPU, batch=128). Shaded regions show the
standard error for 5 trials with different random seeds.}
  \label{fig:ablation_pred_pop}
\end{figure}

Figure \ref{fig:ablation_pred_pop}a highlights that the choice of the underlying predictor algorithm has little impact on the performance of LINAS. Next, in Figure \ref{fig:ablation_pred_pop}b we compare various LINAS runs with different population sizes where a population of 50 gives the best performance for the MobileNetV3 super-network. Finally, we note that while the intent of LINAS is to run for the fewest number of evaluations as possible, an extended run shows that it would take NSGA-II a significant amount of evaluations to catch up with the LINAS hypervolume at 20,000 evaluations. For the subsequent experiments and consistency, we compare LINAS (with NSGA-II for the internal predictor loop) against validation-only measurements from a random search that uniformly samples the architecture space and NSGA-II itself using the algorithm and the predictor settings in Table \ref{tab:search_settings}. 

\begin{figure}[htb]
    \centering
    \includegraphics[width=1.0\linewidth]{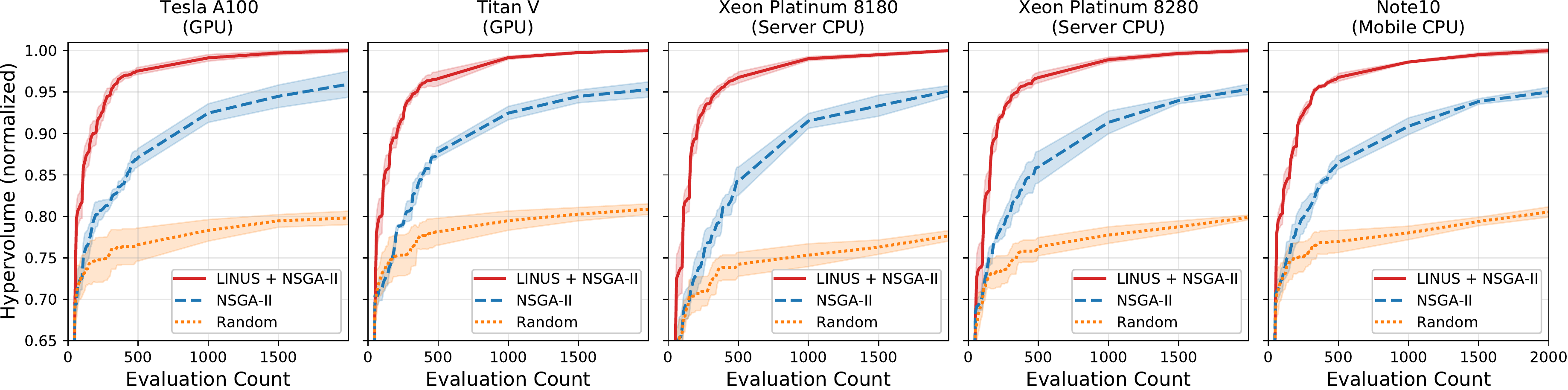}
    \caption{Search comparison on various hardware platforms for the MobileNetV3 super-network.}
    \label{fig:ablation_hardware}
\end{figure}

\begin{figure}[htb]
  \centering
  \begin{subfigure}[t]{0.32\linewidth}
    \centering
    \includegraphics[width=0.98\linewidth]{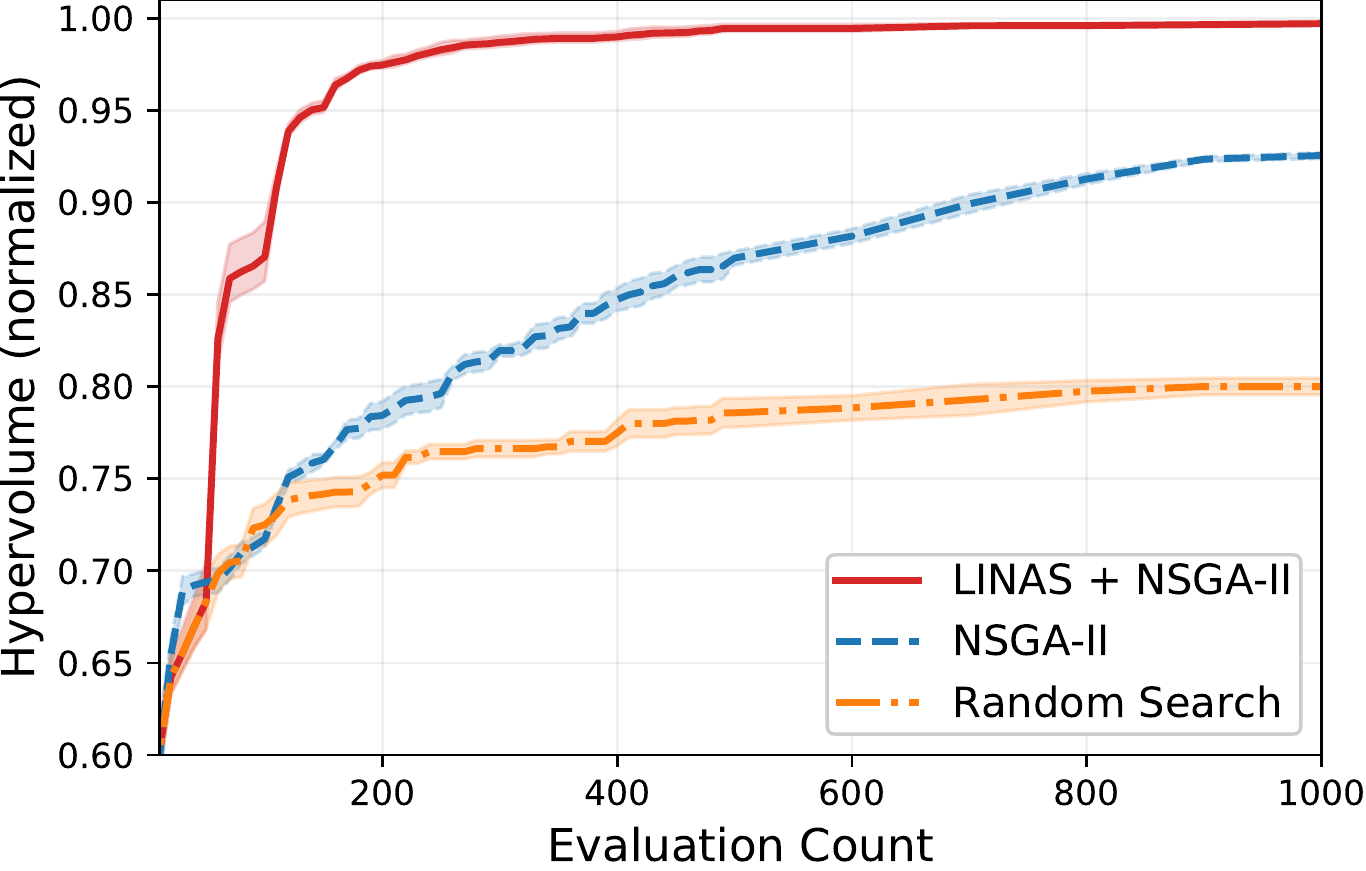}
    \caption{ResNet50}
    \label{hvsubfigure_left}
  \end{subfigure}
  \begin{subfigure}[t]{0.32\linewidth}
    \centering
    \includegraphics[width=0.98\linewidth]{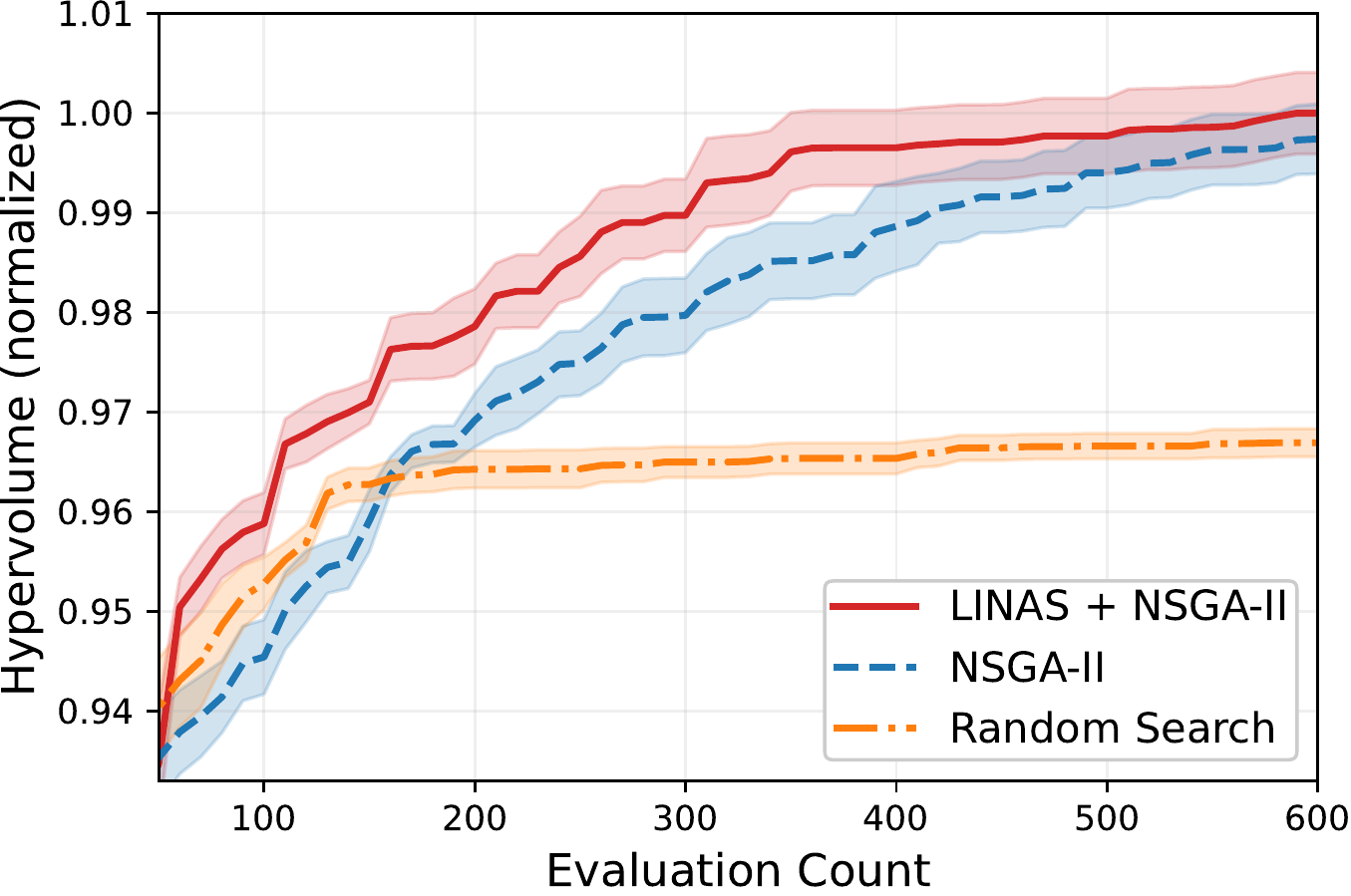}
    \caption{Transformer}
    \label{hvsubfigure_mid}
  \end{subfigure}
  \begin{subfigure}[t]{0.32\linewidth}
    \centering
    \includegraphics[width=0.98\linewidth]{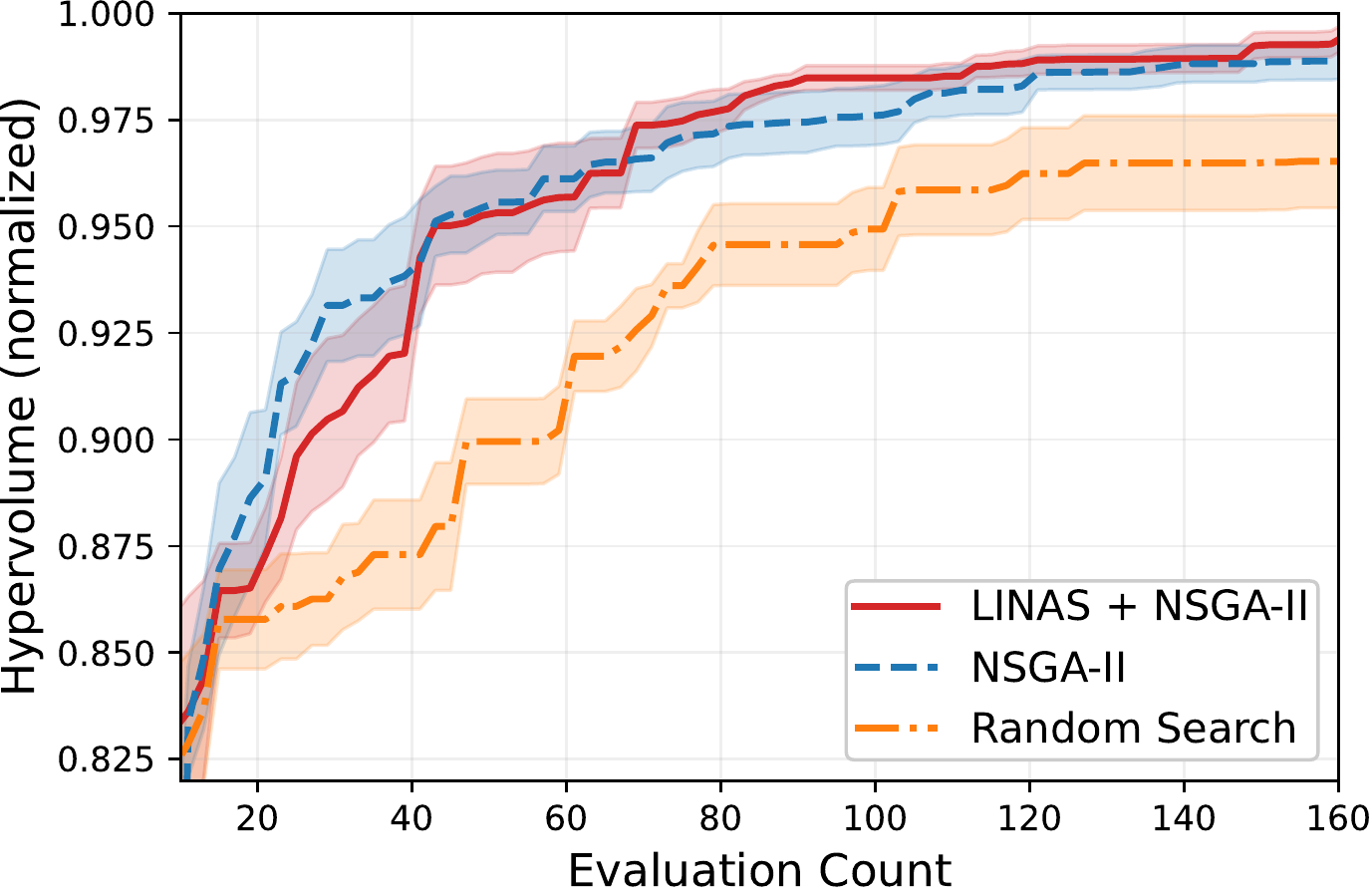}
    \caption{NCF}
    \label{hvsubfigure_right}
  \end{subfigure}
  \caption{Hypervolume comparison for LINAS, NSGA-II, and random search across modalities (Titan-V GPU). Shaded regions = standard error for 5 trials. Search parameter settings in Table \ref{tab:search_settings}.}
  \label{fig:hypervolume_set}
\end{figure}

In this work, we note that each hardware platform has very specific DNN inference time or latency characteristics and a benefit of LINAS is that it can be used to run an accelerated NAS process without any prior knowledge of the latency range. Figure \ref{fig:ablation_hardware} shows a consistent behavior for LINAS across GPU, CPU and mobile hardware settings. When considering the performance of LINAS across various modalities as shown in Figure \ref{fig:hypervolume_set}, a key observation is how quickly LINAS accelerates to a better hypervolume versus the baseline NSGA-II and random search. Depending on which region of the Pareto front is most important, an end-user would be more likely to identify optimal architectures in fewer evaluations with LINAS. Evaluation (validation measurement) counts directly correlate to the search time since the evolutionary algorithm runtime component is far smaller than evaluation runtimes (compute time breakdown given in Appendix \ref{sec:test_platforms}). Given the characteristics of the Transformer and NCF objective spaces, the LINAS result is less differentiated than in the image classification cases. We found that since the distribution of the sub-networks in these super-networks is both more constrained in range and occurs closer to an optimal region, one would be more likely to randomly find a good performing sub-network than in the MobileNetV3 or ResNet50 search spaces. Specifically, NCF is heavily biased towards matrix factorization as described by \citet{rendle2020neural} and we discuss this more in Appendix \ref{sec:appendix_supernets}. Nevertheless, LINAS provides some benefit for all super-network types.

\section{Conclusion}

We have proposed and demonstrated a NAS framework and LINAS algorithm that efficiently finds a \emph{diverse} multi-objective Pareto set of sub-networks from a pre-trained super-network for various modalities. As NAS research continues to gain momentum, we highlight the need to continue to investigate the generalizability of NAS approaches in modalities outside of computer vision. Future work will include extending to experiments to a larger variety of hardware platforms (e.g., TinyML) and application specific deep learning accelerators. Additionally, we plan to explore NAS without training and meta-learning approaches to further reduce the evaluation overhead. 


\bibliographystyle{plainnat}
\bibliography{references}

\clearpage

\appendix

\section{Test Platforms and Compute Time}
\label{sec:test_platforms}

In this work, we use both CPU and GPU platforms for running our experiments. The hardware platforms and their characteristics are shown in Table \ref{tab:hardware_platforms}. For the Note10 mobile CPU experiment shown in Figure \ref{fig:ablation_hardware}, we use a latency look-up table provided by \cite{cai2019once} since we did not have direct access to that platform.

\begin{table*}[htb]
	\caption{Hardware platforms used for NAS experimentation.}
	\centering
	\begin{tabular}{c|c|c|c}
	\hline \hline
	Name & Memory & 
	\begin{tabular}{@{}c@{}} Thread Count \\ (Host CPU)\end{tabular} &
	\begin{tabular}{@{}c@{}} Microarchitecture \\ (Host CPU)\end{tabular} \\
	\hline
	Intel$\textsuperscript{\textregistered}$ Xeon$\textsuperscript{\textregistered}$ Platinum 8180 & 192 GB & 56 & Skylake (SKX) \\
	Intel$\textsuperscript{\textregistered}$ Xeon$\textsuperscript{\textregistered}$ Platinum 8280 & 192 GB & 56 & Cascade Lake (CLX) \\
	NVIDIA$\textsuperscript{\textregistered}$ Titan V$\textsuperscript{\textregistered}$ & 32 GB & 32 & Cascade Lake (CLX) \\
	NVIDIA$\textsuperscript{\textregistered}$ Tesla$\textsuperscript{\textregistered}$ V100 & 32 GB & 32 & Skylake (SKX) \\
	NVIDIA$\textsuperscript{\textregistered}$ Tesla$\textsuperscript{\textregistered}$ A100 & 32 GB  & 32 & Cascade Lake (CLX) \\
	\hline \hline
	\end{tabular}
    \label{tab:hardware_platforms}
\end{table*}


In terms of GPU wall clock time required to perform search, we note that there is a wide range of results that would be dependent on the supporting hardware platform configuration. For example, for MobileNetV3, a sub-network search with 2000 evaluations would take approximately 9.5 GPU hours with an evolutionary algorithm run time on the order of minutes. Because the evolutionary algorithm run times are extremely small when compared to validation measurement run times, we view the evaluation count (e.g., Figure \ref{fig:hypervolume_set}) as a more universal metric of search time efficiency in this work.

To provide more insights into time complexity of presented algorithms, an extensive set of tests was performed to measure wall-clock time of each algorithm needed to achieve a certain hypervolume threshold on each of the search spaces presented in this work. For each super-network, the hypervolume thresholds were selected based on the maximum hypervolume achieved by random search and NSGA-II for a given search space, respectively. In the latter case, the results for random search are not shown as it never achieved the given hypervolume level within a set number of evaluations. Table \ref{tab:algo_time_comparison} shows detailed information on how much time was spent on the model evaluation and the search process itself.

\begin{table}[htb]
\caption{Comparison of algorithms and their average run time on all presented search spaces to a given normalized hypervolume (HV) threshold based on a platform with NVIDIA$\textsuperscript{\textregistered}$ Titan V$\textsuperscript{\textregistered}$ (evaluation) and Intel$\textsuperscript{\textregistered}$ Xeon$\textsuperscript{\textregistered}$ Platinum 8280 (search).}

\centering
\begin{tabular}{|c|ccccc|}
\hline
Super-Network                & \multicolumn{1}{c|}{Search Algorithm} & \multicolumn{1}{c|}{Evaluations} & \multicolumn{1}{c|}{\begin{tabular}[c]{@{}c@{}}Evaluation Cost\\ (GPU Hours)\end{tabular}} & \multicolumn{1}{c|}{\begin{tabular}[c]{@{}c@{}}Search Cost\\ (CPU Hours)\end{tabular}} & \begin{tabular}[c]{@{}c@{}}Total Cost\\ (Hours)\end{tabular} \\ \hline
\multirow{8}{*}{MobileNetV3} & \multicolumn{5}{c|}{Normalized HV = 0.810}                                                                                                                                                                                                                                                                                    \\ \cline{2-6} 
                             & \multicolumn{1}{c|}{LINAS + NSGA-II}  & \multicolumn{1}{c|}{100}         & \multicolumn{1}{c|}{0.479}                                                                 & \multicolumn{1}{c|}{0.0095}                                                            & 0.489                                                        \\ \cline{2-6} 
                             & \multicolumn{1}{c|}{NSGA-II}          & \multicolumn{1}{c|}{260}         & \multicolumn{1}{c|}{1.247}                                                                 & \multicolumn{1}{c|}{0.0014}                                                            & 1.248                                                        \\ \cline{2-6} 
                             & \multicolumn{1}{c|}{Random}           & \multicolumn{1}{c|}{2000}        & \multicolumn{1}{c|}{9.593}                                                                 & \multicolumn{1}{c|}{0.0017}                                                            & 9.594                                                        \\ \cline{2-6} 
                             & \multicolumn{5}{c|}{Normalized HV = 0.955}                                                                                                                                                                                                                                                                                    \\ \cline{2-6} 
                             & \multicolumn{1}{c|}{LINAS + NSGA-II}  & \multicolumn{1}{c|}{346}         & \multicolumn{1}{c|}{1.746}                                                                 & \multicolumn{1}{c|}{0.0331}                                                            & 1.779                                                        \\ \cline{2-6} 
                             & \multicolumn{1}{c|}{NSGA-II}          & \multicolumn{1}{c|}{2000}        & \multicolumn{1}{c|}{9.593}                                                                 & \multicolumn{1}{c|}{0.0017}                                                            & 9.594                                                        \\ \cline{2-6} 
                             & \multicolumn{1}{c|}{Random}           & \multicolumn{1}{c|}{---}         & \multicolumn{1}{c|}{---}                                                                   & \multicolumn{1}{c|}{---}                                                               & ---                                                          \\ \hline
\multirow{8}{*}{ResNet50}    & \multicolumn{5}{c|}{Normalized HV = 0.800}                                                                                                                                                                                                                                                                                    \\ \cline{2-6} 
                             & \multicolumn{1}{c|}{LINAS + NSGA-II}  & \multicolumn{1}{c|}{57}          & \multicolumn{1}{c|}{0.545}                                                                 & \multicolumn{1}{c|}{0.0047}                                                            & 0.549                                                        \\ \cline{2-6} 
                             & \multicolumn{1}{c|}{NSGA-II}          & \multicolumn{1}{c|}{255}         & \multicolumn{1}{c|}{2.438}                                                                 & \multicolumn{1}{c|}{0.0014}                                                            & 2.439                                                        \\ \cline{2-6} 
                             & \multicolumn{1}{c|}{Random}           & \multicolumn{1}{c|}{1000}        & \multicolumn{1}{c|}{9.559}                                                                 & \multicolumn{1}{c|}{0.0016}                                                            & 9.560                                                        \\ \cline{2-6} 
                             & \multicolumn{5}{c|}{Normalized HV = 0.925}                                                                                                                                                                                                                                                                                               \\ \cline{2-6} 
                             & \multicolumn{1}{c|}{LINAS + NSGA-II}  & \multicolumn{1}{c|}{155}         & \multicolumn{1}{c|}{1.481}                                                                 & \multicolumn{1}{c|}{0.0142}                                                            & 1.496                                                        \\ \cline{2-6} 
                             & \multicolumn{1}{c|}{NSGA-II}          & \multicolumn{1}{c|}{1000}        & \multicolumn{1}{c|}{9.559}                                                                 & \multicolumn{1}{c|}{0.0016}                                                            & 9.560                                                        \\ \cline{2-6} 
                             & \multicolumn{1}{c|}{Random}           & \multicolumn{1}{c|}{---}         & \multicolumn{1}{c|}{---}                                                                   & \multicolumn{1}{c|}{---}                                                               & ---                                                          \\ \hline
\multirow{8}{*}{Transformer} & \multicolumn{5}{c|}{Normalized HV = 0.967}                                                                                                                                                                                                                                                                                    \\ \cline{2-6} 
                             & \multicolumn{1}{c|}{LINAS + NSGA-II}  & \multicolumn{1}{c|}{111}         & \multicolumn{1}{c|}{1.886}                                                                 & \multicolumn{1}{c|}{0.0035}                                                            & 1.890                                                        \\ \cline{2-6} 
                             & \multicolumn{1}{c|}{NSGA-II}          & \multicolumn{1}{c|}{191}         & \multicolumn{1}{c|}{3.246}                                                                 & \multicolumn{1}{c|}{0.0014}                                                            & 3.248                                                        \\ \cline{2-6} 
                             & \multicolumn{1}{c|}{Random}           & \multicolumn{1}{c|}{600}         & \multicolumn{1}{c|}{10.197}                                                                & \multicolumn{1}{c|}{0.0015}                                                            & 10.199                                                       \\ \cline{2-6} 
                             & \multicolumn{5}{c|}{Normalized HV = 0.997}                                                                                                                                                                                                                                                                                               \\ \cline{2-6} 
                             & \multicolumn{1}{c|}{LINAS + NSGA-II}  & \multicolumn{1}{c|}{465}         & \multicolumn{1}{c|}{7.903}                                                                 & \multicolumn{1}{c|}{0.0156}                                                            & 7.918                                                        \\ \cline{2-6} 
                             & \multicolumn{1}{c|}{NSGA-II}          & \multicolumn{1}{c|}{600}         & \multicolumn{1}{c|}{10.197}                                                                & \multicolumn{1}{c|}{0.0015}                                                            & 10.199                                                       \\ \cline{2-6} 
                             & \multicolumn{1}{c|}{Random}           & \multicolumn{1}{c|}{---}         & \multicolumn{1}{c|}{---}                                                                   & \multicolumn{1}{c|}{---}                                                               & ---                                                          \\ \hline
\multirow{8}{*}{NCF}         & \multicolumn{5}{c|}{Normalized HV = 0.965}                                                                                                                                                                                                                                                                                    \\ \cline{2-6} 
                             & \multicolumn{1}{c|}{LINAS + NSGA-II}  & \multicolumn{1}{c|}{87}          & \multicolumn{1}{c|}{2.884}                                                                 & \multicolumn{1}{c|}{0.0138}                                                            & 2.898                                                        \\ \cline{2-6} 
                             & \multicolumn{1}{c|}{NSGA-II}          & \multicolumn{1}{c|}{89}          & \multicolumn{1}{c|}{2.950}                                                                 & \multicolumn{1}{c|}{0.0014}                                                            & 2.952                                                        \\ \cline{2-6} 
                             & \multicolumn{1}{c|}{Random}           & \multicolumn{1}{c|}{160}         & \multicolumn{1}{c|}{5.304}                                                                 & \multicolumn{1}{c|}{0.0014}                                                            & 5.305                                                        \\ \cline{2-6} 
                             & \multicolumn{5}{c|}{Normalized HV = 0.989}                                                                                                                                                                                                                                                                                    \\ \cline{2-6} 
                             & \multicolumn{1}{c|}{LINAS + NSGA-II}  & \multicolumn{1}{c|}{148}         & \multicolumn{1}{c|}{4.906}                                                                 & \multicolumn{1}{c|}{0.0241}                                                            & 4.930                                                        \\ \cline{2-6} 
                             & \multicolumn{1}{c|}{NSGA-II}          & \multicolumn{1}{c|}{160}         & \multicolumn{1}{c|}{5.304}                                                                 & \multicolumn{1}{c|}{0.0014}                                                            & 5.305                                                        \\ \cline{2-6} 
                             & \multicolumn{1}{c|}{Random}           & \multicolumn{1}{c|}{---}         & \multicolumn{1}{c|}{---}                                                                   & \multicolumn{1}{c|}{---}                                                               & ---                                                          \\ \hline
\end{tabular}
\label{tab:algo_time_comparison}
\end{table}

\section{Hardware Platform Transferability}
\label{sec:hardware_platform_transfer}

\begin{figure}[htb]
    \centering
    \includegraphics[width=0.6\linewidth]{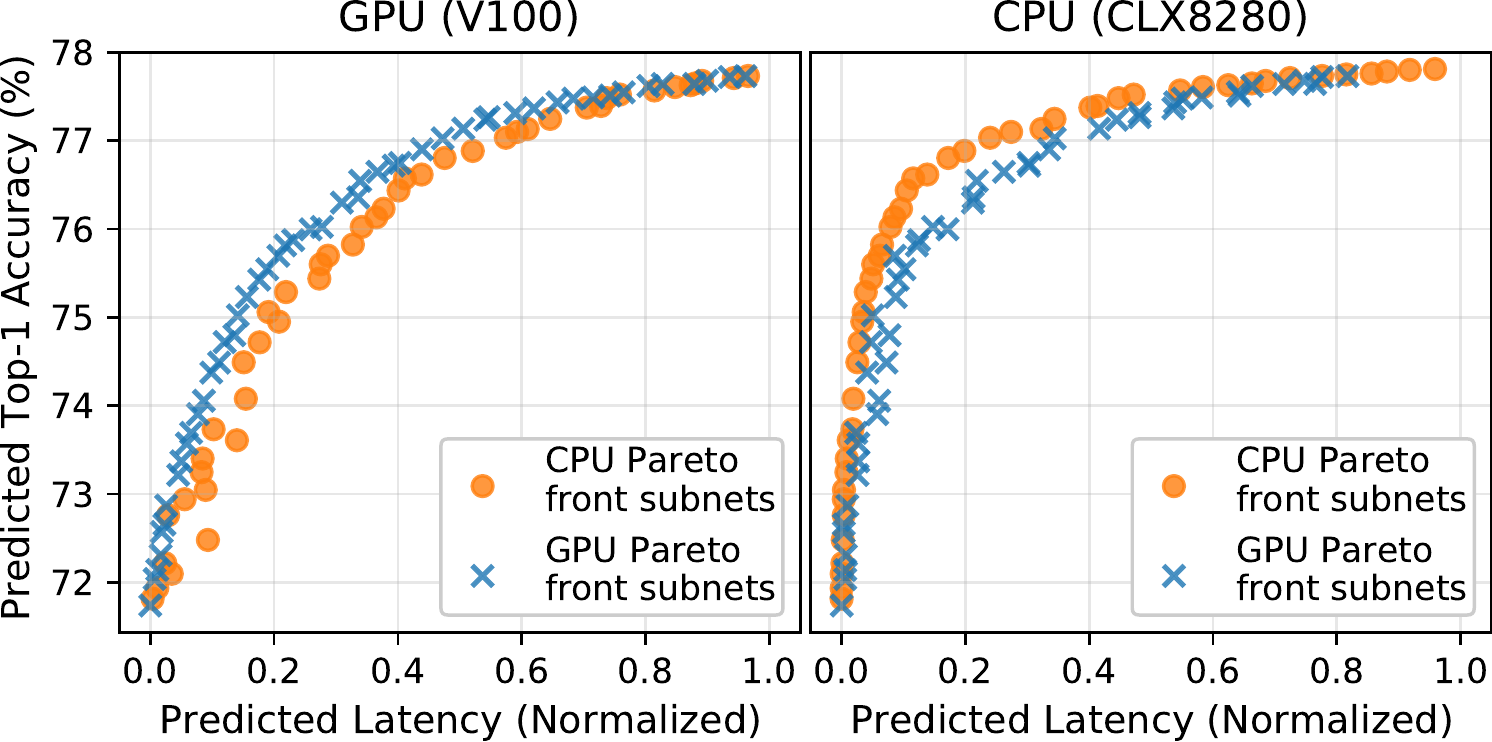}
    \caption{MobileNetV3 Pareto fronts specialized to GPU (V100) and CPU (CLX) showing that optimal sub-network configurations found on one hardware platform do not translate to the optimal sub-networks for another. Batch size was 128. }
    \label{fig:pareto_hw_compare}
\end{figure}

\begin{figure}[htb]
    \centering
    \includegraphics[width=1.0\linewidth]{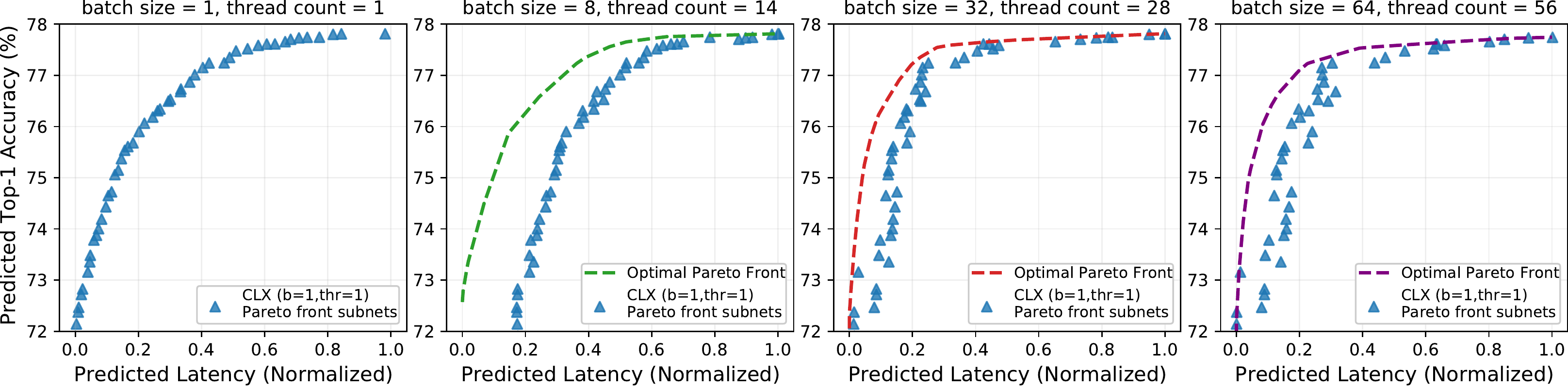}
    \caption{MobileNetV3 Pareto fronts with CLX for specialized thread counts/batch sizes, and the non-specialized configurations for comparison.}
    \label{fig:pareto_clx_configs}
\end{figure}

One of the key goals of our framework is to accelerate the sub-network search process to address the issue that every hardware platform and/or configuration has unique latency characteristics and therefore unique optimal sub-networks in their respective multi-objective search spaces. To illustrate this behavior, we use the MobileNetV3 super-network where Figure \ref{fig:pareto_hw_compare} shows that an optimal set of sub-networks found on a CPU platform may not transfer to the optimal objective region on a GPU platform and vice versa. Furthermore, within a hardware platform, Figure \ref{fig:pareto_clx_configs} shows that sub-network configurations found to be optimal to one CPU hardware configuration (e.g., batch size = 1, thread count = 1), do not transfer optimally to other hardware batch size and thread count configurations. 

\section{Latency Prediction}
\label{sec:latency_prediction}

\begin{figure}[htb]
    \centering
    \includegraphics[width=1.0\linewidth]{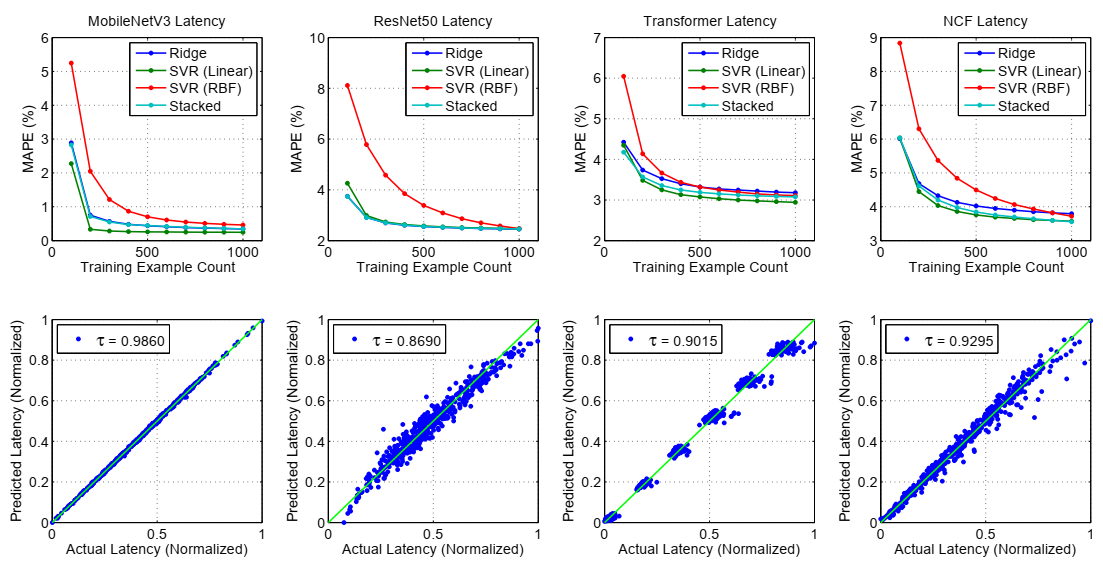}
    \caption{MAPE of predictors performing latency prediction versus the number of training examples for sub-networks derived from the super-networks shown in Table \ref{tab:super_network_summary} (top row). Correlation and Kendall $\tau$ coefficient between actual and predicted latencies after training the stacked predictor with 1000 examples (bottom row). The ideal correlation is shown by the green line.}
    \label{fig:latency_predictor_analysis}
\end{figure}

The analysis of latency prediction is performed in the same way as described in Section \ref{ssec:accuracy_predictor_analysis} with the results shown in Figure \ref{fig:latency_predictor_analysis}. The top row shows the MAPE of different predictors for each super-network type. The bottom row shows the correlation between actual and predicted latencies after training the stacked predictor with 1000 examples. Note that the Kendall rank correlation coefficient $\tau$ is also shown for each case.

\section{Super-network Details}
\label{sec:appendix_supernets}

\subsection{MobileNetV3}
\label{sec:mobilenetv3}

For the image classification task with MobileNetV3, we experiment on the ImageNet validation dataset \citep{imagenet} and use the pre-trained super-network weights from  \emph{ofa\_mbv3\_d234\_e346\_k357\_w1.0}\footnote{\label{note_ofa}https://github.com/mit-han-lab/once-for-all}, trained with progressive shrinking. For the architecture design variables, we allow for an elastic layer depth chosen from [2, 3, 4], an elastic width expansion ratio chosen from [3, 4, 6], an elastic kernel size chosen from [3, 5, 7], and use an input image resolution of 224x224. The layer depth can affect the mapping of the kernel size and expansion ratio design variables as shown in Figure \ref{fig:encoding}(a). For more details on this super-network please refer to the work by \citet{cai2019once}.

\subsection{ResNet50}
\label{sec:resnet50-OFA}

For the image classification task with ResNet50, we use the ImageNet validation dataset \citep{imagenet} and use the pre-trained super-network weights from  \emph{ofa\_resnet50\_d=0+1+2\_e=0.2+0.25+0.35\_w=0.65+0.8+1.0}\textsuperscript{\ref{note_ofa}}, trained with progressive shrinking for our experiments. For the architecture design variables, we allow for an elastic layer depth chosen from [0, 1, 2], an elastic width expansion ratio chosen from [0.65, 0.8, 1.0], an elastic expansion ratio chosen from [0.2, 0.25, 0.35], and use an input image resolution of 224x224.

\subsection{Transformer}
For the machine translation task, we mainly experiment on the WMT 2014 En-De data set. We follow a similar pre-processing technique proposed in \citep{wang2020hat} for the data. Similar to \citep{wang2020hat}, we use the  search space with an embedding dimension chosen from [512, 640] , hidden dimension from [1024, 2048, 3072] , attention head number from [4, 8], decoder layer number from [1, 2, 3, 4, 5, 6] and a constant encoder layer number of [6]. In \citep{wang2020hat}, although the authors use the inherited weights from the Transformer super-network, for the evolutionary search, they re-train the sub-networks from scratch in the final results. In our results, we do not re-train these networks from scratch. Additionally, we train the predictor directly on the \emph{bilingual evaluation understudy (BLEU)} \citep{papinenibleu2002} score. For the BLEU score evaluation, we use a beam size of 5 and a length penalty of 0.6. 

\subsection{NCF}
For the recommendation task, we experiment on the Pinterest-20 dataset and follow a similar pre-processing technique used in \citep{he2017neural} . We use the \textit{Neural Matrix Factorization} (NeuMF) model from \citep{he2017neural}, which is a fusion of \textit{Generalized Matrix Factorization} (GMF) and \textit{Multi-Layer Perceptron} (MLP).   We create an elastic NCF super-network model with the embedding dimension for MLP and GMF layers sampled from [8, 16,32,64, 128], MLP layer number from [1,2,3,4,5,6], and MLP hidden sizes from [8, 16, 32, 64, 128, 256, 512, 1024]. We train the NCF super-network by uniformly sampling different sub-networks for each mini-batch of training.

In our experiments we see that when the matrix factorization module of the NCF sub-network was sufficiently large, the results of the subnetwork were dominated by it versus the MLP module. \citet{rendle2020neural} substantiate this hypothesis in their paper by showing that a well-tuned matrix factorization approach can substantially outperform proposed learned similarities such as an MLP. We thus attribute the diminished improvement in the performance of LINAS on NCF to the degeneracies in the search space caused by a more powerful matrix factorization module which strongly dominates the HR@10.

\section{Multiply-Accumulates to Latency}
\label{sec:appendix_macs}

\begin{figure}[htb]
  \centering
  \begin{subfigure}[t]{0.45\linewidth}
    \centering
    \includegraphics[width=0.9\linewidth]{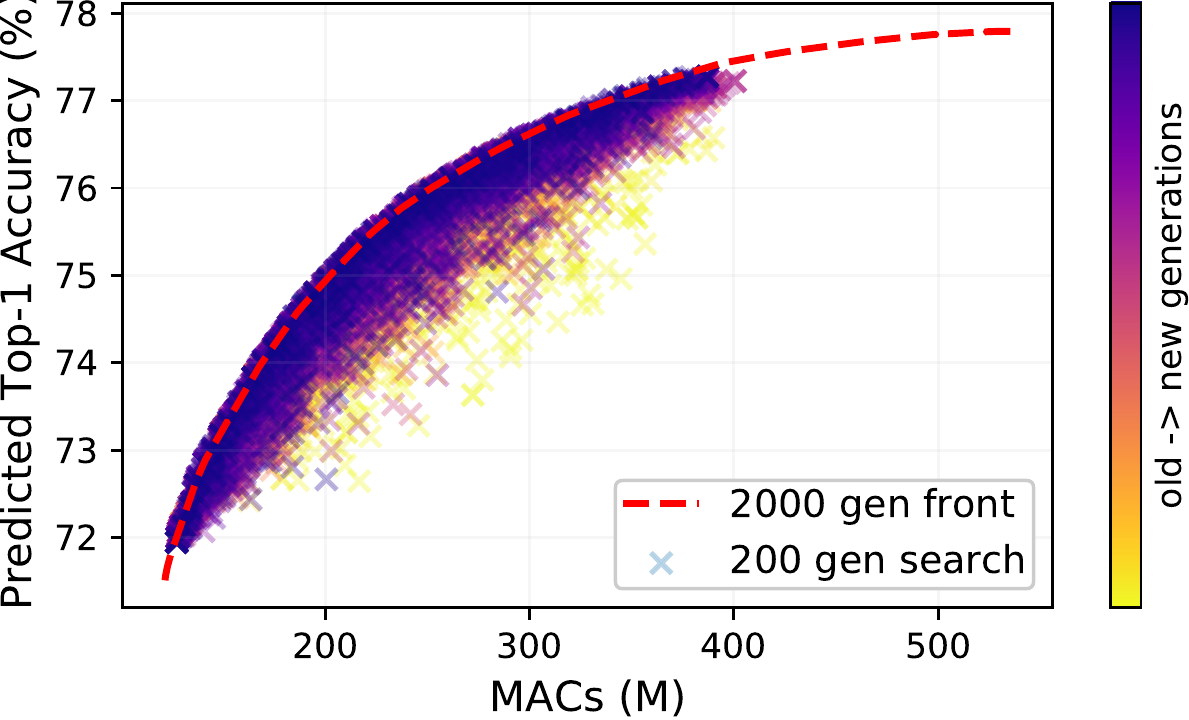}
    \caption{NSGA-II search progression in the MACs versus top-1 accuracy objective space.}
    \label{sfig:search_macs}
  \end{subfigure}
  \hspace{5mm}
  \begin{subfigure}[t]{0.45\linewidth}
    \centering
    \includegraphics[width=0.9\linewidth]{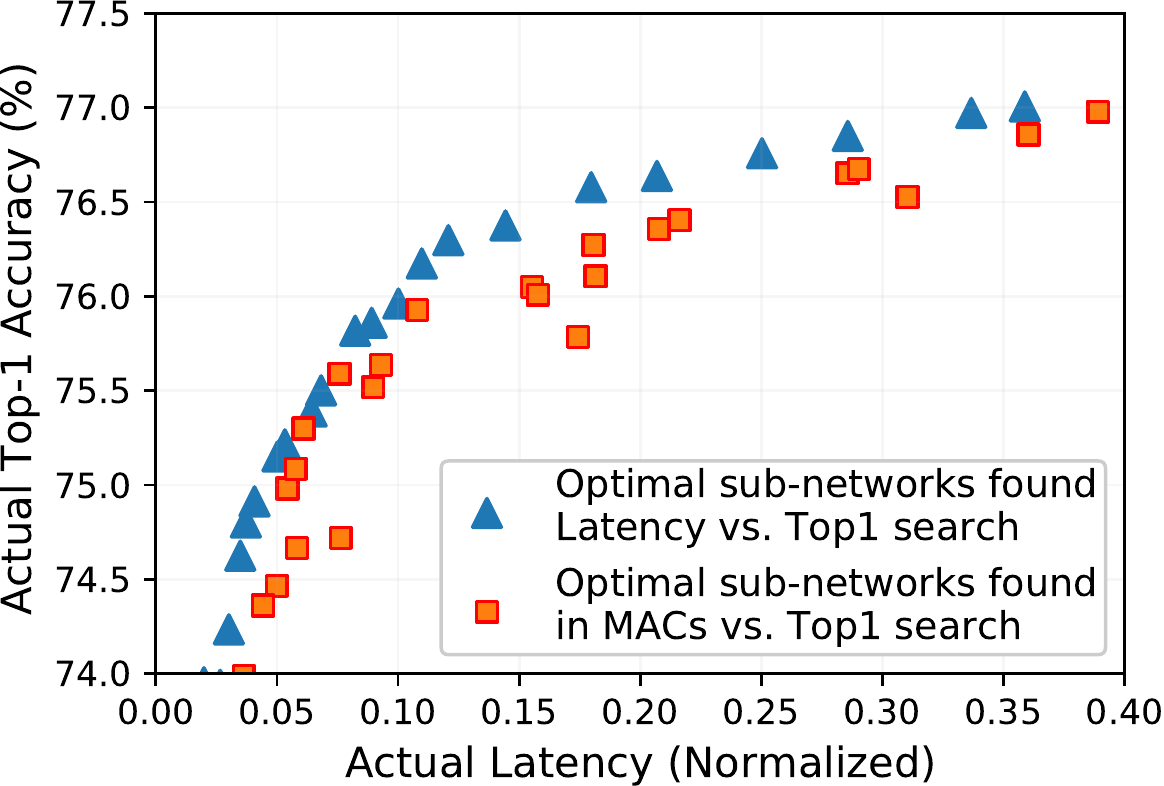}
    \caption{MACs-based optimal sub-networks placed in the latency space. CLX CPU with batch=128, threads=56.}
    \label{sfig:compare_macs}
  \end{subfigure}
  \caption{Comparison between latency and MACs-based NSGA-II searches using the MobileNetV3 super-network showing that the best Pareto front sub-networks from a MACs-based search do not always translate optimally to the latency objective space.}
  \label{fig:compare_macs_latency}
\end{figure}

In addition to evaluating search performance on the latency, accuracy (Top-1), and BLEU score objectives, we looked at the search trends in terms of multiply-accumulates (MACs) and accuracy as shown in Figure \ref{sfig:search_macs} using the fvcore\footnote{https://github.com/facebookresearch/fvcore} library.  Often, multiply-accumulates (MACs) or floating point operations per second (FLOPs) are used to approximate latency. However, we note that the transferrability between these metrics has its limitations. For example, Figure \ref{sfig:compare_macs} highlights that optimal sub-networks identified during a lengthy (e.g., run search until the Pareto front is saturated with sub-network options) multi-objective MACs and top-1 accuracy NSGA-II search do not translate to the most optimal sub-networks identified during a latency-based NSGA-II search. One benefit of a MACs search is that the best Pareto front population would be ideal for a warm-start population on subsequent searches for a given super-network. Another option in our framework would be to perform a many-objective search (e.g., U-NSGA-III) to find optimal sub-networks in the latency, accuracy, \emph{and} MACs search space. 

\section{LINAS Performance for Quantized Super-Networks}
\label{sec:bnas_search}

In addition to searching for optimal configurations using the OFA image classification super-networks (based in FP32 number format), we also experimented with finding optimal INT8 models from the novel BootstrapNAS (BNAS) ResNet50 super-network \citep{bootstrapNAS}. BNAS transforms a single reference pre-trained DNN architecture into a super-network and streamlines the sub-network search process in the quantized INT8 space. Specifically, we leverage a BNAS ResNet50 super-network (BNAS-ResNet50Q) that has a different design space, specified with different values for the search design variables, than the Once-for-all ResNet50 model discussed in Section \ref{sec:resnet50-OFA}. In this setup, the elastic layer depth chosen from [0, 1], an elastic width expansion ratio chosen from [0.65, 0.8, 1.0] and an elastic expansion ratio chosen from [0.2, 0.25].

\begin{figure}[htb]
    \centering
    \includegraphics[width=0.5\linewidth]{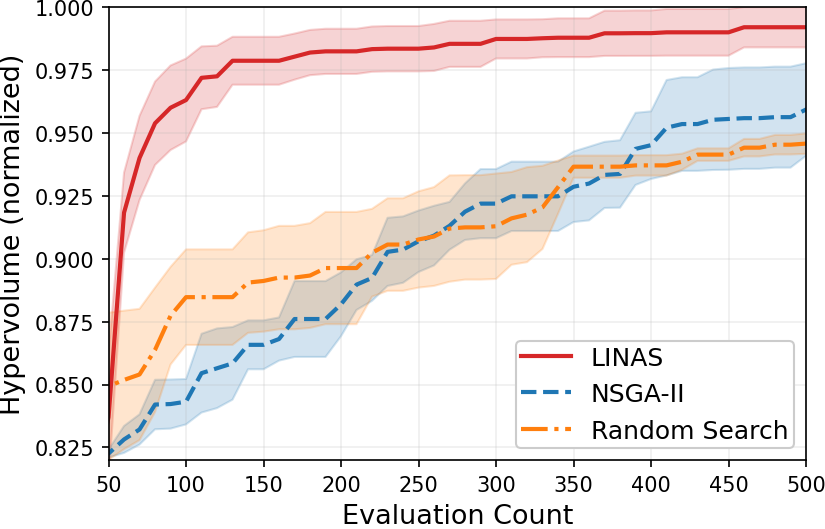}
    \caption{Hypervolume comparison of LINAS, NSGA-II and Random Sampling search methods applied to the quantized model (INT8) of BootstrapNAS ResNet50 super-network. Shaded regions show the standard error for 5 trials with different random seeds.}
    \label{fig:hypervolume_bnas}
\end{figure}

\begin{figure}[htb]
  \centering
  \begin{subfigure}[t]{0.49\linewidth}
    \centering
    \includegraphics[width=1.0\linewidth]{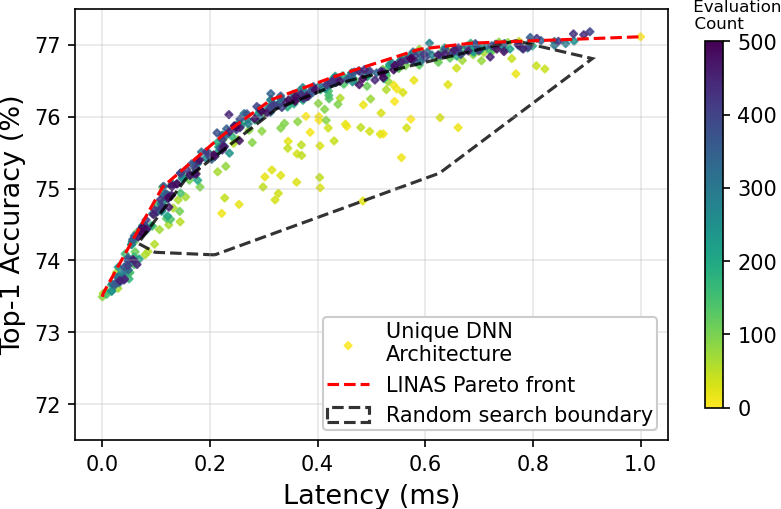}
    \caption{LINAS Search}
    \label{subfigure_linas}
  \end{subfigure}
  \begin{subfigure}[t]{0.49\linewidth}
    \centering
    \includegraphics[width=1.0\linewidth]{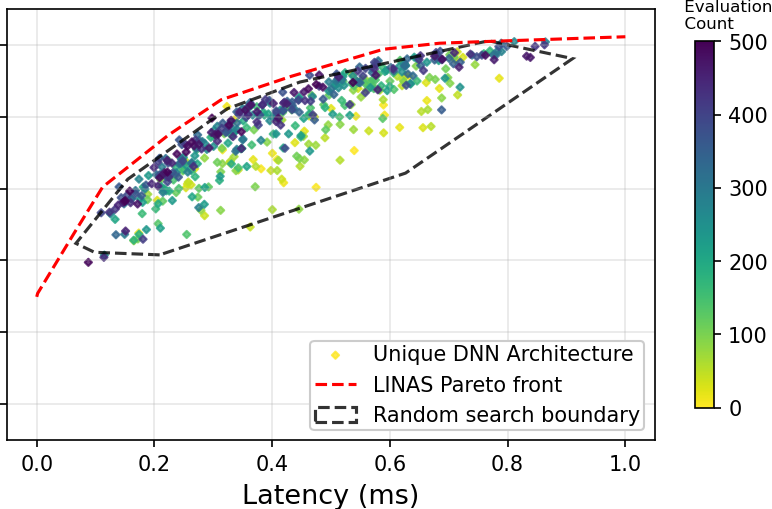}
    \caption{NSGA-II Search}
    \label{subfigure_nsga}
  \end{subfigure}
  \hfill
  \caption{Search results in the BootstrapNAS ResNet50 INT8 search space (CLX, batch size = 128) comparing \subref{subfigure_linas}: LINAS, and \subref{subfigure_nsga}: NSGA-II approaches.}
  \label{fig:bnas_scatter_plots}
\end{figure}

Our experiment follows the steps outlined in Figure \ref{fig:system_flow} with an additional weight conversion from FP32 to INT8 and standardized fine-tuning of the INT8 model in the last step. As shown on the Figure \ref{fig:hypervolume_bnas}, LINAS offers improvements in terms of hypervolume progression and in the time required to find diverse models in the Pareto front. NSGA-II under-performs in early stages of the search when compared to random search, which could be explained by the characteristics of the BootstrapNAS super-network, and its smaller selection of elastic parameters that have been limited to those promising better performance for the extracted sub-networks. As shown on the Figure \ref{fig:bnas_scatter_plots}, the overall distribution of the randomly sampled configurations is closer to the optimal region of the objective space, which may be the cause of random search yielding comparable results in the early stages of the search process. This result is likely due to the nature of BNAS' process for selecting promising elastic design parameters and also that the search space derived by BNAS-ResNet50Q ($\sim10^{7}$) is much smaller than OFA's ResNet-50 ($\sim10^{13}$). 

\section{Search Algorithm Details}
\label{sec:search_algorithm_settings}

For the search algorithm comparison study in Section \ref{ssec:linas_results}, we evaluate the performance of various evolutionary algorithms, a SMBO multi-objective tree-structured parzen estimator (MOTPE), and a random search using the MobileNetV3 super-network. The evolutionary algorithm settings used for the experiments are shown in Table \ref{tab:ea_tournament_settings}. Evolutionary algorithms that support two or more objectives typically fall in the categories of indicator- or decomposition-based algorithms where the latter often use a predefined set of reference directions on a unit simplex to create objective space partitions. For generating a well-spaced set of reference points from the objective space origin we use the Riesz s-Energy approach \citep{refdirenergy}. For the MOTPE parameters, we use the recommended settings provided by the authors \citet{optuna_2019} including a prior weight of 1.0 and number of candidate samples used to calculate the expected hypervolume improvement equal to 24.

\begin{table}[tb]
\caption{Evolutionary algorithm parameter settings for the comparison study in Figure \ref{fig:ea_tournament}. Settings generally follow those recommended by \citet{pymoo} for each algorithm. }
\centering
\begin{tabular}{cccccc}
\hline
\hline
\multicolumn{1}{l}{}                                                         & \multicolumn{5}{c}{Evolutionary Algorithm}                                                                                                                                                                                                                \\ \cline{2-6} 
                                                                             & NSGA-II & AGE-MOEA & U-NSGA-III                                                               & C-TAEA                                                                   & MOEA/D                                                                   \\ \hline
\begin{tabular}[c]{@{}c@{}}Number of \\ supported \\ objectives\end{tabular} & 2       & 2        & $\geq$ 2                                                               & $\geq$ 2                                                                & $\geq$ 2                                       \\ \hline
\begin{tabular}[c]{@{}c@{}}Population\\ size\end{tabular}               & 50    & 50     & 50                                                                     & -                                                                    & -                           
\\ \hline
\begin{tabular}[c]{@{}c@{}}Mutation\\ probability\end{tabular}               & 0.02    & 0.02     & 0.02                                                                     & 0.05                                                                     & -                                                                        \\ \hline
\begin{tabular}[c]{@{}c@{}}Crossover \\ probability\end{tabular}             & 0.9     & 0.9      & 0.9                                                                      & 1.0                                                                      & -                                                                        \\ \hline
\begin{tabular}[c]{@{}c@{}}Reference \\ direction\\ method\end{tabular}      & -       & -        & \begin{tabular}[c]{@{}c@{}}Riesz s-Energy\\ (20 partitions)\end{tabular} & \begin{tabular}[c]{@{}c@{}}Riesz s-Energy\\ (20 partitions)\end{tabular} & \begin{tabular}[c]{@{}c@{}}Riesz s-Energy\\ (20 partitions)\end{tabular} \\ \hline
\begin{tabular}[c]{@{}c@{}}Number of \\ neighbors\end{tabular}               & -       & -        & -                                                                        & -                                                                        & 20                                                                       \\ \hline
\begin{tabular}[c]{@{}c@{}}Neighbor \\ mating\\ probability\end{tabular}     & -       & -        & -                                                                        & -                                                                        & 0.9                                                                      \\ \hline
\end{tabular}
\label{tab:ea_tournament_settings}
\end{table}

\begin{table}[htb]
\caption{Experiment settings for the LINAS (with NSGA-II internal loop) and NSGA-II comparison studies in Figures \ref{fig:mbnv3_scatter_plots}, \ref{fig:ablation_pred_pop}, \ref{fig:ablation_hardware} and \ref{fig:hypervolume_set}. The predictor types apply only to the LINAS setup.}
\centering
\begin{tabular}{c|ccc}
\hline
\hline
\begin{tabular}[c]{@{}c@{}}Super-Network\\ (Modality)\end{tabular} & \begin{tabular}[c]{@{}c@{}}Transformer\\ (Machine Translation)\end{tabular} & \begin{tabular}[c]{@{}c@{}}MobileNetV3, ResNet50\\ (Image Classification)\end{tabular} &
\begin{tabular}[c]{@{}c@{}}NCF\\ (Recommendation)\end{tabular} \\ \hline

Accuracy Predictor                                                 & SVR w/ RBF kernel                                                                        & Ridge  & SVR w/ RBF kernel                                                              \\
Latency Predictor                                                 & Ridge                                                                        & Ridge  & SVR w/ Linear kernel \\
Search Space                                                      & $10^{15}$                                                                   & $10^{19}$ & $10^{7}$                                                           \\
Population                                                        & 50                                                                          & 50  & 10                                                                  \\
Crossover                                                         & 0.9                                                                         & 0.9      & 0.1                                                            \\
Mutation                                                          & 0.02                                                                        & 0.02   & 0.02  \\
\begin{tabular}[c]{@{}c@{}}LINAS evaluations \\(Predictor)\end{tabular} & 20000 & 20000 & 2000 \\ \hline
\end{tabular}
\label{tab:search_settings}
\end{table}

For the LINAS and NSGA-II experiments across different modalities (Figures \ref{fig:mbnv3_scatter_plots}, \ref{fig:ablation_pred_pop}, \ref{fig:ablation_hardware} and \ref{fig:hypervolume_set}) we show parameter settings in Table \ref{tab:search_settings}. In our ablation studies, cross-over rates between 0.9 and 1.0 performed nearly the same, smaller populations work well with smaller search space sizes, and a mutation rate equal to the inverse of the population size as recommended by \citep{pymoo} gives the best search performance for NSGA-II. The same settings were used for the LINAS inner-loop predictor search which also often uses NSGA-II in this work. An important note is that NSGA-II and AGE-MOEA are not compatible with three or more objectives and that the other many-objective EA approaches, such as U-NSGA-III would need to be considered in that setting.

\end{document}